\documentclass[twoside,11pt]{article}
\usepackage{jmlr2e}
\usepackage[utf8x]{inputenc}
\usepackage{dirtytalk}
\usepackage{comment}
\usepackage[linesnumbered, ruled, vlined]{algorithm2e}
\usepackage{algorithmic}
\usepackage{float}
\usepackage{placeins}
\usepackage{multirow}
\usepackage{xcolor}

\newcommand{\methodACRON}{COMPER}
\newcommand{\method}{\textit{\textbf{COMP}act \textbf{E}xperience \textbf{R}eplay}}

\ShortHeadings{Improving Experience Replay through Modeling of Similar Transitions' Sets}
{Neves, Batisteli, Lopes, Ishitani \& Patrocínio Jr}
\firstpageno{1}

\usepackage{caption}
\usepackage{subcaption}
\usepackage{multirow}

\begin{document}

\title{Improving Experience Replay through\\ Modeling of Similar Transitions' Sets}

\author{\name Daniel Eugênio Neves \email daniel.eugenio@sga.pucminas.br \\
        \name João Pedro Oliveira Batisteli \email joao.batisteli@sga.pucminas.br\\
        \name Eduardo Felipe Lopes \email eflopes@sga.pucminas.br\\
        \name Lucila Ishitani \email lucila@pucminas.br\\
        \name Zenilton Kleber Gonçalves do Patrocínio Júnior \email zenilton@pucminas.br\\
        \addr Pontifícia Universidade Católica de Minas Gerais,\\
       Rua Walter Ianni, 255,  São Gabriel, Belo Horizonte - MG - CEP: 31980110}
       

\thispagestyle{plain}

\maketitle
\begin{abstract}%

In this work, we propose and evaluate a new reinforcement learning method, \textit{\textbf{COM}Pact \textbf{E}xperience \textbf{R}eplay} (COMPER), which uses temporal difference learning with predicted target values based on recurrence over sets of similar transitions, and a new approach for experience replay based on two transitions memories. Our objective is to reduce the required number of experiences to agent training regarding the total accumulated rewarding in the long run. Its relevance to reinforcement learning is related to the small number of observations that it needs to achieve results similar to that obtained by relevant methods in the literature, that generally demand millions of video frames to train an agent on the Atari 2600 games. We report detailed results from five training trials of COMPER for just 100,000 frames and about 25,000 iterations with a small experiences memory on eight challenging games of Arcade Learning Environment (ALE). We also present results for a DQN agent with the same experimental protocol on the same games set as the baseline.  To verify the performance of COMPER on approximating a good policy from a smaller number of observations, we also compare its results with that obtained from millions of frames presented on the benchmark of ALE.

\end{abstract}
\section{Introduction}\label{sec:Introduction}

Machine learning is an artificial intelligence research field in which mathematical and statistical models applied through computational algorithms seek to provide machines with intelligent behavior and the ability to learn from their experiences. In particular, reinforcement learning (RL) research dedicates to the development of computational entities called \say{agents}, which try to learn the best way to interact with their environment to achieve a given goal. Each action of an agent affects the environment. Then the current state is observed by it along with a reward signal representing the effects of those actions. Thus, the reward values allow it to assess how its action was (that is, how good its behavior was) in the sense of keeping it closer or farther from its goal. In this context, the total expected long-term reward provides a measurement of goal fulfillment by an agent~\citep{Sutton2018}.

Relevant research works in literature present proposals based on refinements of psychology and neuroscience concepts that provided the theoretical bases for the development of reinforcement learning. Some of these works explore methods of temporal difference learning, which unites trial-and-error processes with optimal control techniques, such as Q-Learning~\citep{Watkins1992}, also combining some kind of memory with repetition of agent experiences~\citep{Lin1992} in approximate models through deep neural networks~\citep{mnih2015humanlevel,DBLP:journals/corr/SchaulQAS15,NIPS2010_3964}. There are many issues and limitations concerning reinforcement learning methods, due to critical factors that emerge as the addressed scenarios become more complex. Among these factors is the required time for agent training, which is directly related to the agent capacity of converging to a better policy of actions. Since the demand for long training times can limit the application of reinforcement learning methods, how to effectively reduce this time is an important question to be answered. 

After a reinforcement learning agent has performed a sequence of actions and received a reward value, knowing how to assign credit or discredit to each state-action pair (to adjust its decision making and improve its effectiveness) consists of a difficult problem, called \textit{temporal credit assignment}~\citep{Lin1992}. Since it is related to the agent effectiveness, more efficient approaches for that problem can also reduce the time spent in its learning process. In this sense, temporal difference (TD) learning represents one of the main techniques to deal with credit attribution, despite being a slow process, especially when it involves credit propagation over a long sequence of actions. As a consequence, Adaptive Heuristic Critic -- AHC~\citep{Sutton92reinforcementlearning} and Q-Learning~\citep{Watkins1992}, which represent the first TD-based methods to address reinforcement learning difficulties, are characterized by high convergence times. To address that issue, \citet{Lin1992} proposed and analyzed a technique named Experience Replay (ER), demonstrating its effectiveness to accelerate the credit attribution process and, consequently, to reduce the convergence time. One of the main motivations for ER proposal is the fact that the traditional algorithms, based on AHC and Q-Learning, become inefficient when trial-and-error experiences are used only once to adjust the models and then discarded. For \citet{Lin1992}, this is a waste, considering that some experiences can be rare, while others can be expensive to obtain, such as those involving penalties.

Experience Replay has some limitations. But it is more efficient in the credit propagation if a sequence of experiences is repeated in the opposite temporal order to the one in which they were experienced. Also, this process can be more efficient if TD($\lambda$)-methods are used with $\lambda>0$, in which $\lambda$ is the number of calculation steps. In this way, the temporal difference error used to adjust models is obtained by the discrepancy between multiple consecutive predictions and not just two~\citep{Lin1992}. Another issue is related to the fact that ER repeats experiences through sampling. In this case, if those sampled experiences define policies that are very different from the one that is being learned, that can lead to an underestimation of the evaluation and utility functions. For ER to be useful, an agent must repeat only experiences involving actions that will continue to follow the policy that is being learned by it. That occurs both for methods based on AHC and Q-Learning. In the case of Q-Learning, this issue affects methods that use approximate models through neural networks because whenever network weights are adjusted for a given state, this affects the entire model concerning many (or perhaps all other) states. Thus, if this network is trained using bad experiences many times in a row, it will tend to underestimate the real usefulness of state-action pairs. However, to prevent overfitting, an agent needs to implement a good strategy to select which experiences should be remembered and how often to repeat them~\citep{Lin1992}.

Different approaches in the literature use ER, and discussions about the process of using memories to repeat experiences have relevance and implications in all of them \citep{DBLP:journals/corr/MnihKSGAWR13,mnih2015humanlevel,van_Hasselt_Guez_Silver_2016,DBLP:journals/corr/SchaulQAS15,Wang:2016:DNA:3045390.3045601}. Among the issues commonly pointed out is the need to better explore experiences repetition to obtain more significant events during sampling. There is also a concern that agents with online reinforcement learning update their parameters based on a sequence of experiences and, in their most simplistic form, discard those experiences immediately after. Since those updates may be strongly correlated, this goes against the assumption that model variables are independent and identically distributed -- i.i.d. (which is adopted by most stochastic gradient methods). Thus, there is no possibility to take advantage of rare experiences that could be useful in the future. Experience Replay addresses these issues because with the experiences stored and sampled from memory it is possible to break the correlation among consecutive experiences. This is done by interleaving those that are more recent with ones that are less recent to be used in model updating while making it possible that rare experiences can be used in the future. Another problem is the fact that the experience memory is size limited. Therefore, some experiences are never repeated before being discarded, while others are repeated for the first time only long after being stored.


Experience Replay is a method of fundamental importance for several reinforcement learning algorithms, but it still presents many questions that have not yet been exhausted and problems that are still open, mainly those related to the use of experiences that can contribute more to accelerate the agent's learning. According to \citet{DBLP:journals/corr/SchaulQAS15}, the use of a memory of experiences presents two main issues: (i)~the selection of experiences to store; and (ii)~the establishment of which ones to repeat. They addressed the latter, assuming that the content of the memory was beyond their control. In turn, we address the first and propose a new method named \method\ (\methodACRON), which explores the way that experience memory can be modeled to make the agents with ER more efficient in using smaller amounts of data. The objective of \methodACRON\ is to reduce the number of experiences needed during its training related to the total accumulated reward expected in the long run. Its relevance to reinforcement learning is related to the small number of observations (video frames) that it needs to achieve results that are similar to the ones obtained by important methods in the literature, such as DQN, Double DQN, and PER, which also are based on temporal difference learning with experience replay. Those methods usually need millions of video frames to train their agents to make good choices on the Atari 2600 games, while \methodACRON\ requires a considerably smaller amount to obtain equivalent results.

The process of training an agent with TD learning may lead to the repetition of transitions with different values (Q-values) for the action-value function over consecutive training episodes (each episode starts in the initial game state and finish when the game is over at a final state, so the game restart and another episode is carried, and so on). In the proposed method, we demonstrate that it is possible to produce sets of similar transitions and establish a relationship between them by indexing these sets. We can explore that index to produce a transitions memory that stores only a single representation of the similar transitions and perform successive updates of the Q-values associated with them, learning the set's dynamics on time through a recurrent neural network. Moreover, we have also demonstrated that it is possible to obtain a smaller memory of transitions, but capable of representing all of them and even so be effective for a model updating based on the temporal difference error. Sampling transitions on that smaller memory augments the chances of a rare transition to be observed compared with a sampling performed on the entire set of already performed transitions and can make the update of the value function even more effective. Finally, we demonstrate that our proposals can contribute to reducing the convergence time of reinforcement learning methods that use temporal difference learning regarding the number of observations required for the agent training. That seems to happen especially on problems where the agents take long times to be rewarded and, therefore, need to learn a long-term action policy based on their experiences.

Thus, \methodACRON\ is a reinforcement learning method based on TD-learning, with Q-Value function approximation, Q-Target value prediction based on recurrence over sets of similar transitions, and a new approach for experience replay based on two transitions memories. Its main features are how it structures and uses its transition memories and its model for predicting the target value. Experimental results show that it achieves pretty good results on complex games of Atari 2600 using considerably fewer video frames than other methods in the literature. In addition, this work brings a new discussion not only about how to deal with the retrieval of an agent's experiences but also about how one can store them.

The rest of this paper is organized as follows. In Section~\ref{sec:literature_review}, some related works are presented. In Section~\ref{sec:background}, we review some concepts related to reinforcement learning and DQN method. After that, our new proposed method is described in Section~\ref{sec:comper}. Section~\ref{sec:metodologia} presents and discuss our experimental methodology, while test results are shown and analyzed in Section~\ref{results}. Finally, we draw some conclusions and possible future works in Section~\ref{sec:conclusion}.
\section{Literature Review and Related Work}\label{sec:literature_review}

Seeking to reduce the convergence time of reinforcement learning methods based on temporal difference learning, \citet{Lin1992} investigated three techniques to accelerate the credit attribution process (i.e., to shorten the trial-and-error process): (i)~Experience Replay; (ii)~Learning Action Models for Planning; and (iii)~Teaching. These were applied in eight reinforcement learning frameworks based on AHC and Q-Learning and their extensions: (i)~AHCON (Connectionist AHC-learning); (ii)~AHCON-R (AHCON using experience replay); (iii)~AHCON-M (AHCON using actions model); (iv)~AHCON-T (AHCON using experience replay and teaching); (v)~QCON (Connectionist Q-Learning); (vi)~QCON-R (QCON with experience replay); (vii)~QCON-M (QCON with actions model); and (viii)~QCON-T (QCON with experience replay and teaching). Those frameworks seek to learn a policy evaluation function to predict the discounted accumulated return to be received by an agent. The evaluation function is modeled using neural networks and trained using a combination of TD and backpropagation. The author stated that: (i) experience replay is an effective technique to speed up the credit assignment process without the need of a provided perfect model of the world or a good model that is quickly learned; (ii) the advantage of using teaching only becomes more significant as the task becomes harder, but that still lacks a teacher; (iii) the superiority of experience replay over actions model when agents need to learn a model of the environment by themselves. For \citet{Lin1992}, experience replay performs a form of planning through sampling past experiences. That is similar to the approach with an action model, but without the need to learn this model. Thus, the main feature of experience replay is to repeat only actions that are under the policy being learned, and this is a very efficient method to speed up the credit assignment process, in addition to being relatively simple to understand, analyze, and implement.


\citet{DBLP:journals/corr/MnihKSGAWR13}, \citet{mnih2015humanlevel}, \citet{DBLP:journals/corr/SchaulQAS15}, and \citet{van_Hasselt_Guez_Silver_2016} proposed and evaluated important methods based on Q-Learning~\citep{Watkins1992} and Double Q-Learning~\citep{NIPS2010_3964} for the development of reinforcement learning agents with the goal of obtaining the highest scores in Atari 2600 games. Their experiments were carried out in the Arcade Learning Environment (ALE) \citep{bellemare13arcade,machado18arcade}, which allows complex challenges for reinforcement learning agents. To approximate the value functions the authors used convolutional neural networks (CNN) on state representations based on the video frames. The networks did not have any prior information regarding the games, no manually extracted features, and no knowledge regarding the internal state of the ALE emulator. Thus, agent learning occurred only from video inputs, reward signals, the set of possible actions, and the final state information of each game. These works adopted ER, but there were important differences concerning selected methods to approximate functions and sampling experiences.

\citet{DBLP:journals/corr/MnihKSGAWR13} proposed the method named Deep Q-Networks (DQN) and demonstrate how their approach achieved state-of-the-art results and human-level performance on Atari 2600 games. DQN was improved by \citet{mnih2015humanlevel} through a change in the way CNN are used. In~\citet{DBLP:journals/corr/MnihKSGAWR13}, the same network (with the same parameters $\Theta$) is used for approximating both the action-value function $Q(s, a, \Theta)$ and the target action-value function $Q(s',a, \Theta)$. In turn, \citet{mnih2015humanlevel}  used independent sets of parameters $\Theta$  and $\Theta'$ for each network. Thus, only the function $Q(s, a, \Theta)$ has its parameters $\Theta$ updated by backpropagation. The parameters $\Theta'$ are updated directly (i.e., copied) from the values of $\Theta$ with a certain frequency, remaining unchanged between two consecutive updates. Thus, only the \textit{forward pass} is carried out, when the network is used with the parameters $\Theta'$ to predict the value of the target function. The authors attributed their state-of-the-art results mainly to the ability of their CNN to represent the games' states. They also pointed to the fact that games that require long-term planning strategies, such as \textit{Montezuma's Revenge}, still represent a major challenge for all reinforcement learning agents.

According to \citet{DBLP:journals/corr/MnihKSGAWR13}, the use of ER with random sampling reduces the effect of the correlation between the data (that represents the environment states) and also improves its non-stationary distribution (during the training of the neural network) because it softens the distribution over many previous experiences. For ~\citet{mnih2015humanlevel}, the correlation between consecutive observations leads to the minor updates on the approximate model to cause considerable changes in the policy learned by the agent. This way, it can change the data distribution and the relation between the action-value functions during the calculation of the TD-error used in the Q-Learning update step. Two limitations in their approaches are: (i) their memories of experiences did not differentiate relevant experiences because of the uniform distribution; and (ii) experiences are often overwritten due to the buffer size limitation. That points to the need for more sophisticated strategies that can emphasize experiences capable of contributing more to agent learning, in the sense of what was proposed by \citet{DBLP:journals/corr/SchaulQAS15}. It is also interesting to note the strong bias imposed by the adoption of maximization along with the target function of Q-Learning. That bias had already motivated the proposition of another method by~\citet{NIPS2010_3964}, named Double Q-Learning, whose formulation was able to tackle this problem as demonstrated by~\citet{van_Hasselt_Guez_Silver_2016}.

For \citet{DBLP:journals/corr/SchaulQAS15}, the use of ER with uniform sampling does not consider the relevance of the experiences for the agent learning, usually repeating them with the same frequency they were experienced. In this sense, the authors proposed the method Prioritized  Experience Replay (PER), which prioritizes some experiences and repeats that with a frequency according to the TD-error magnitude, based on the idea that an agent can learn more efficiently from some experiences than others, and that some of them may not be immediately relevant to the agent, but can become as it learns. They adopted a stochastic prioritization process and an importance-based sampling method to mitigate the fact that prioritization can lead to a diversity loss and some bias insertion.

For \citet{DBLP:journals/corr/HausknechtS15}, mapping states to actions based only on the four previous game states (stacking the frames in a preprocessing step) prevents the DQN agent from achieving the best performance in games that require remembering faraway events from a large number of frames, because in those games the future states and rewards depend on several previous states. Therefore, the authors proposed using a long short-term memory (LSTM) in place of the first fully connected layer, just after the series of convolutive layers in the original architecture of DQN, to get better use of the limited history. The authors demonstrate that there is a trade-off between using a non-recurrent network with a long history of observations or a recurrent network with just one frame at each iteration step. They stated that a recurrent network is a viable approach for dealing with observations from multiple states, but it presents no systematic benefits compared to stacking these observations in the input layer of a plain CNN. \citet{moreno2019performing} proposed a similar approach but using DDQN instead of DQN.

\citet{Wang:2016:DNA:3045390.3045601} proposed an architecture named Dueling Network in which they used two parallel streams (instead of a single sequence of fully connected layers) just after the convolutional layers, that are combined in the output by an aggregation layer to produce the estimates of Q-values. Based on their results, they claimed to have achieved significant improvements in agent performance compared to DQN and DDQN and achieved the state-of-the-art with the use of PER.

\citet{hambow_2018} investigated how to combine DDQN, PER, Dueling Network, Asynchronous Advantage Actor-Critic (A3C)~\citep{pmlr-v48-mniha16}, Distributional Q-Learning~\citep{pmlr-v70-bellemare17a}, and Noisy DQN~\citep{DBLP:journals/corr/FortunatoAPMOGM17} and integrated these different but complementary ideas into an approach named Rainbow since each one of them contributes with significant improvements in different aspects. The authors stated that their results provided the state-of-the-art performance on ALE~\citep{bellemare13arcade} for 57 games and presented an ablation study about the contributions of each component to the overall agent performance, following the evaluation procedures of ~\citet{mnih2015humanlevel} and~\citet{van_Hasselt_Guez_Silver_2016}.

\citet{DBLP:journals/corr/abs-1712-01275} presented an empirical study about Experience Replay that demonstrated that a large replay buffer can harm the agent performance and verified that the replay buffer length is a very important hyperparameter that has been neglected in the literature. They proposed a method to minimize the negative influence of a large replay buffer, named Combined Experience Replay (CER), consisting of adding the last transition to the sampled batch before using it in agent training. They hypothesized that there is a trade-off between the data quality and the data correlation as smaller replay buffers make data fresher but highly temporal correlated, while neural networks often need i.i.d. data. In turn, data sampling from larger replay buffers tends to be uncorrelated but is outdated. If the replay buffer is full and works as a queue (i.e. FIFO) this will affect the agent's learning. But if we assume that different transitions from many stochastic episodes carried out over a non-deterministic environment will be stored and sampled many times from a smaller replay buffer but not explicitly size limited, we can hypothesize that this buffer tends to be less correlated along with the time while the buffer size increases. 

\citet{DBLP:journals/corr/abs-1903-00374} presented a model-based deep reinforcement learning algorithm with a video prediction model named SimPLe, which performed well after just 102,400 interactions (that correspond to 409,600 frames on ALE and about 800,000 samples from the video prediction model) and compared their results with the ones obtained by Rainbow~\citep{hambow_2018}. Their objective was to show that planning with a parametric model allows for data-efficient learning on several Atari video games. In that sense,~\citet{NEURIPS2019_1b742ae2} proposed a broad discussion about model-based algorithms and experience replay pointing out its commonalities and differences, when to expect benefits from either approach and how to interpret prior works in this context. They set up experiments in a way comparable to~\citet{DBLP:journals/corr/abs-1903-00374} and demonstrated that in a like-for-like comparison, Rainbow outperformed the scores of the model-based agent, with less experience and computation. Rainbow used a total number of 3.2 million replayed samples, and SimPLe used 15.2 million.~\citet{kaiser2020modelbased} presented their final published paper comparing SimPLe and Rainbow on the number of iterations needed to achieve the best results. SimPLe achieved the best game scores on half of the game set. Although~\citet{machado18arcade} recommends evaluating the agent performs on the final scores of different checkpoints over training episodes, \citet{kaiser2020modelbased} states that one of the SimPLe limitations is that its final scores are on the whole lower than the best state-of-the-art model-free methods. We discuss in detail our methodology for experiments and agent evaluation in Section~\ref{sec:metodologia}.

Experience Replay~\citep{Lin1992} was a fundamental idea to reinforcement learning and is still being investigated by many researchers to understand its contributions and propose improvements~\citep{NIPS2017_453fadbd,horgan2018distributed,ijcai2018-666,pmlr-v97-novati19a,ijcai2019-589,pmlr-v119-fedus20a,wei2021deep,daley2021stratified}. In this work, we present a method that addresses the experience memory and how we can model and explore it to make the agents with ER more efficient in using smaller amounts of data.

\section{Background}\label{sec:background}

Reinforcement learning is learning what to do---how to map situations to actions---so as to maximize a numerical reward signal~\citep{Sutton2018}. It uses the formal framework of Markov Decision Processes (MDPs) to define the interaction between a learning agent and its environment in terms of states,
actions, and rewards.

A MDP is a tuple $(\cal{S}, \cal{A}, \cal{P}, R, \gamma)$, in which $\cal{S}$ represents a set of states, $\cal{A}$ is a set of actions $\cal{A}$$=\{a_1,a_2,\ldots,a_n\}$, and $\cal{P}$ is a state probability function. In this way, ${\cal{P}}(s'\,|\,s, a)$ is the probability for transiting from state $s$ to $s′$ by taking action $a$, $\cal{R}$ is a reward function mapping each state-action pair to a reward in $\mathbb{R}$, and $\gamma \in [0, 1]$ is a discount factor. The agent behavior is represented by a policy $\pi$ and the value $\pi(a\,|\, s)$ represents the probability of taking action $a$ at state $s$. 
At each time step $t$, an agent interacting with the MDP, observes a state $s_t \in \cal{S}$,
and chooses an action $a_t \in \cal{A}$ which determines the reward $r_t = {\cal{R}}(s_t, a_t)$ and next state $s_{t+1} \sim {\cal{P}}(\cdot\,|\,s_t, a_t)$.
A discounted sum of future rewards is called return $R_t = \sum_{t'=t}^{\infty} \gamma^{t'-t} r_{t'}$. Its goal is to learn (or approximate) an optimal policy $\pi^\ast$ that maximizes the long-term expected (discounted) return value.

Q-Learning~\citep{Watkins1992} is a model-free off-policy algorithm that applies successive steps to update the estimates for the action-value function $Q(s, a)$ (that approximate the long-term expected return of executing an action from a given state) using temporal difference learning and minimizing the temporal-difference error (defined by the difference on Equation ~\ref{eq:q-learing}). This function is named Q-function and its estimated returns are known as Q-values. A higher Q-value indicates that an action $a$ is judged to yield better long-term results in state $s$.


Q-Learning converges to an $\pi^\ast$ even if it is not acting optimally every time as long it keeps updating all state-action. Under this assumption and with a variation of the usual stochastic approximation conditions, which generates a sequence of variations in the value of the \(\alpha\) variable, the Q-function converges to an optimal action-value function~\citep{Sutton2018}, as we can describe in Equation~\ref{eq:q-learing} :

\begin{equation}
    Q(s_t,a_t) \leftarrow Q(s_t,a_t)+\alpha [r +\gamma \max_aQ(s_{t+1},a)- Q(s_t,a_t)].
    \label{eq:q-learing}
\end{equation}

\begin{figure}[t]
\centering
\includegraphics[width=0.99\textwidth]{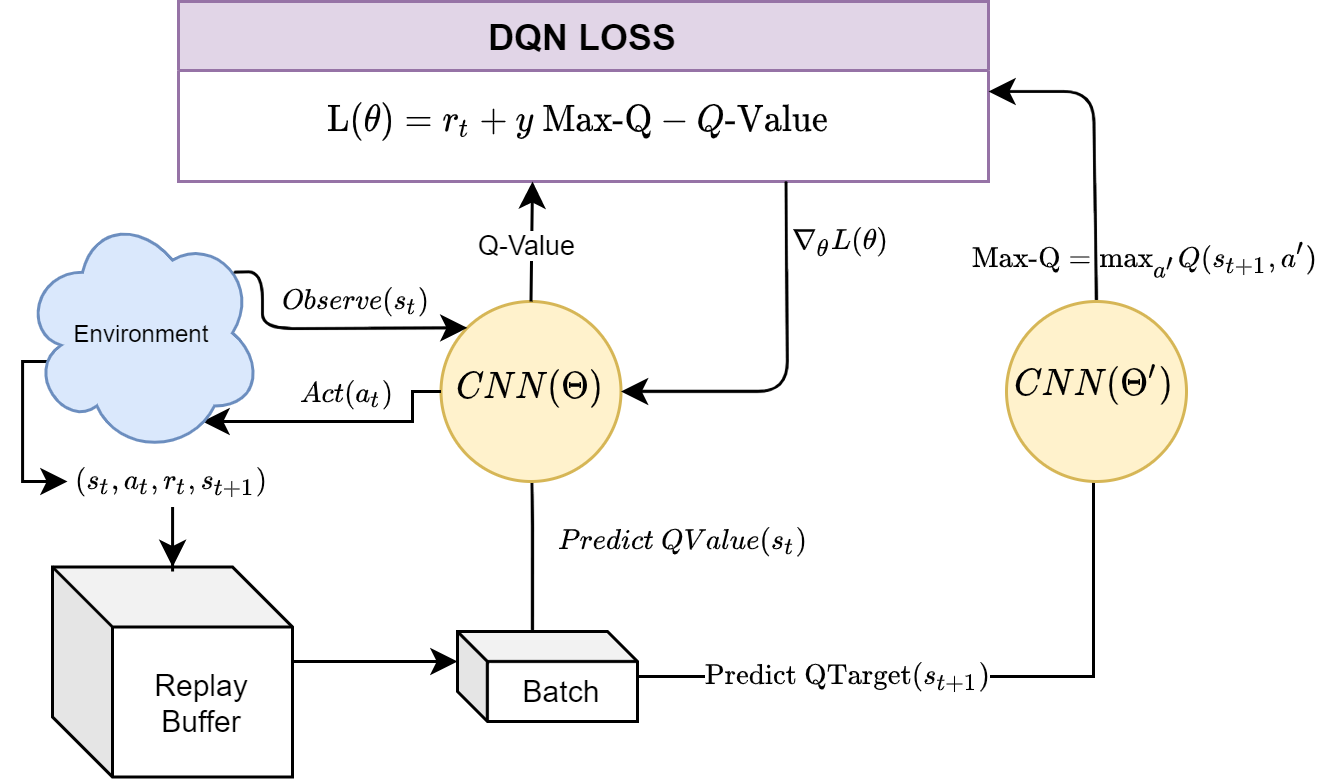}
\caption{General DQN flow and componentes.}
\label{fig:methods_flows-DQN}
\end{figure}

The DQN algorithm~\citep{mnih2015humanlevel} is based on Q-Learning and uses a deep neural network to approximate the Q-function, which is parameterized by $\Theta$ as $Q(s, a, \Theta)$. 
Moreover, to train that neural network in DQN, a technique known as Experience Replay~\citep{Lin1992} is adopted to break the strong correlations between consecutive state inputs during the learning. Specifically, at each time-step $t$, a transition (or experience) is defined by a tuple $\tau_t = (s_t, a_t, r_t, s_{t+1})$, in which $s_t$ is the current state, $a_t$ is the action taken at that state, $r_t$ is the received reward at $t$, and $s_{t+1}$ is the resulting state after taking action $a_t$. 
Recent experiences are stored to construct an experience replay buffer ${\cal{D}} = \{\tau_1, \tau_2,\ldots, \tau_{N_{\cal{D}}}\}$, in which $N_{\cal{D}}$ is the buffer size. 
Therefore, a deep neural network can be trained on samples of experiences
$(s_t, a_t, r_t, s_{t+1}) \sim U({\cal{D}})$, drawn uniformly at random from the pool of stored samples by iteratively minimizing the following loss function,
\begin{equation}
{\cal{L}}_{DQN}(\Theta_i)= {\mathbb{E}}_{(s_t, a_t, r_t, s_{t+1}) \sim U({\cal{D}})} \left[ \left( r_t + \gamma \max_{a'} \widehat{Q}(s_{t+1}, a′, \Theta') - Q(s_t, a_t, \Theta_i) \right)^2 \right],
\end{equation}
in which $\Theta_i$ are the parameters from the $i$-th iteration.
Instead of using the main network, another one provides update values $\widehat{Q}(s_{t+1}, a′, \Theta')$ utilized to obtain target values for the temporal difference error, decoupling any feedback that may result from using the network to generate its own targets.
In~\citet{mnih2015humanlevel}, the target network architecture was identical to the main one except for its parameters $\Theta'$ that are updated to match $\Theta$ after a fixed number of iterations. According to~\citet{tsitsiklis1997analysis}, the use of the same network to generate the next state target Q-values that are used in updating
its current Q-values may lead to oscillation and even divergent behavior, which represents the motivation to the adoption of techniques to restore learning stability, such as  sampling from a replay buffer and using a different target network (even if it shares the same architecture). Figure~\ref{fig:methods_flows-DQN} illustrates DQN algorithm.

In the Atari 2600 games domain, it has been shown that DQN is able to learn the Q-function with pixel-based low-level inputs in an end-to-end manner~\citep{mnih2015humanlevel}.
Learning is performed by sampling experiences from the replay memory to update the network parameters, instead of using online data in the original order. One should observe that target values depend on the network weights, which is in contrast with supervised learning (in which targets are fixed before learning begins). Finally, to balance exploration and exploitation, given an estimated Q-function, DQN adopts the $\epsilon$-greedy strategy to generate the experiences.

\FloatBarrier

Generally, a convolutional neural network (CNN) is used to approximate both the estimated Q-values and the target values. 
Some approaches in the literature have proposed the adoption of recurrent units (and even some sort of attention mechanism) at the final layers of the target network~\citep{DBLP:journals/corr/HausknechtS15,moreno2019performing}. However they have found out that their proposals present no systematic benefits compared to the adoption of \textit{frame stacking} with a CNN without any recurrent unit.


\section{\method\ (\methodACRON) }\label{sec:comper}

In this section, we present the proposed method named \method\ (\methodACRON) along with its theoretical hypotheses and formal mathematical definition.

\begin{figure}[t]
\centering
\includegraphics[width=0.99\textwidth]{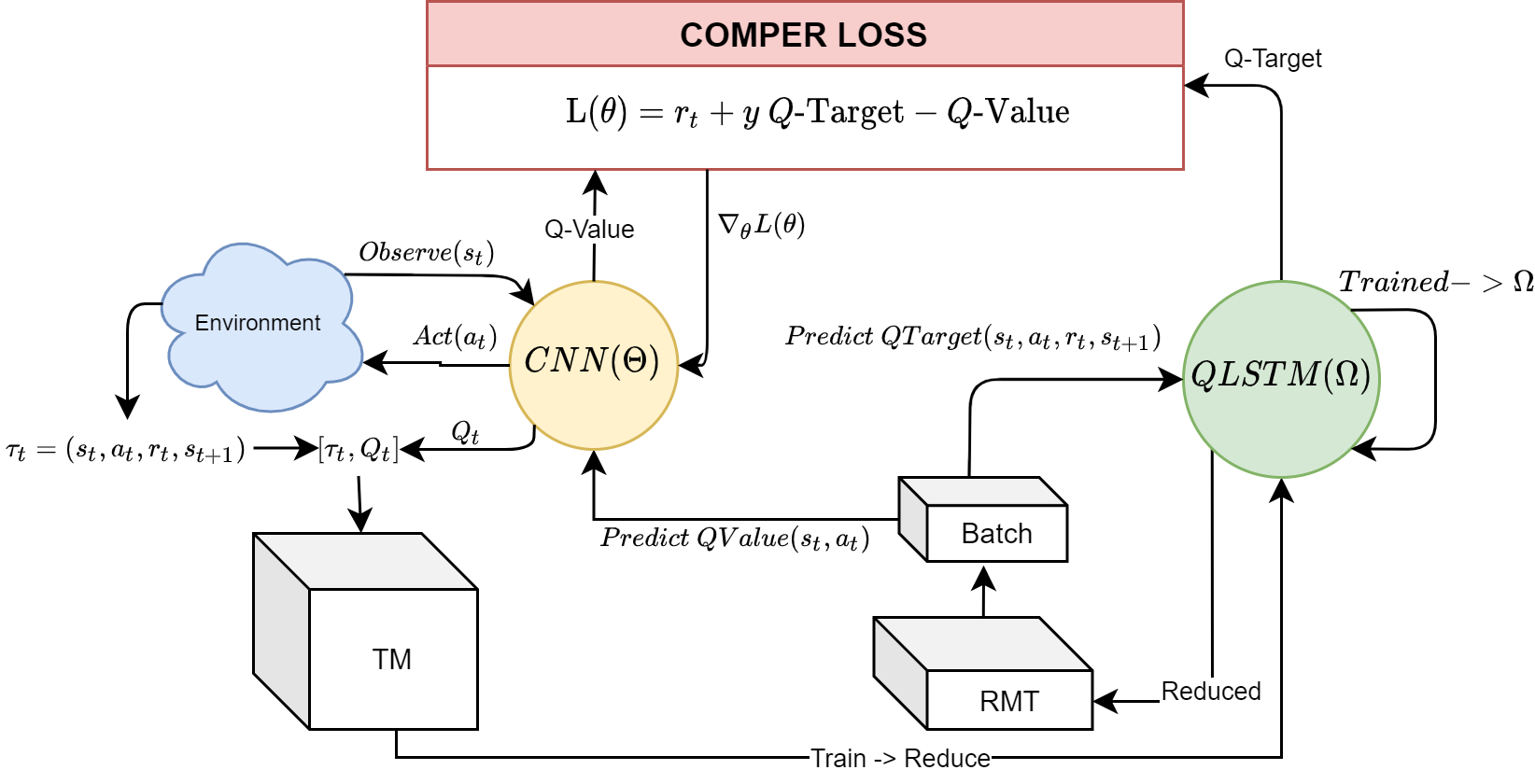}
\caption{General COMPER flow and componentes.}
\label{fig:methods_flows-COMPER}
\end{figure}

\subsection{Method Outline}


Similar to DQN, the proposed method \methodACRON\ also adopts experience replay and updates $Q(s,a)$ using temporal-difference error. However, it does not construct just an experience replay buffer. Instead, \methodACRON\ samples transitions from a much more compact structure named Reduced Transition Memory (${\cal{RTM}}$). The goals behind that are two-fold: (i) allowing rare (and possibly expensive) experiences to be sampled more frequently; and (ii) decreasing the amount of memory needed by a large experience buffer without hampering the convergence time. To achieve that, \methodACRON\ first stores transitions together with estimated Q-function values into a structure hereafter named Transition Memory (${\cal{TM}}$) -- this is similar to a traditional experience replay buffer, except for the presence of the Q-value and the identification and indexing of Similar Transitions Sets (${\cal{ST}}$). 

After that, we can explore the similarities between transitions (explained in Section~\ref{sec:transitions_memory}) stored in ${\cal{TM}}$ to generate a more compact (but still very helpful) version called ${\cal{RTM}}$. Thus, we can use this for transitions sampling during training a deep neural network. The transitions
$(s_t, a_t, r_t, s_{t+1}) \sim U({\cal{RTM}})$, are drawn uniformly at random from the ${\cal{RTM}}$ by iteratively minimizing the following loss function,

\begin{equation}
{\cal{L}}_{COMPER}(\Theta_i)= {\mathbb{E}}_{\tau_t=(s_t, a_t, r_t, s_{t+1}) \sim U({\cal{RTM}})} \left[ \left( r_t + \gamma\, R(\tau_t, \Omega) - Q(s_t, a_t, \Theta_i) \right)^2 \right],
\end{equation}
\noindent in which $Q(s_t, a_t, \Theta_i)$ is an approximation of Q-function made by a deep neural network parameterized by $\Theta_i$ at $i$-th iteration. 
In \methodACRON, update values $R(\tau_t, \Omega)$ used to obtain target values are provided by a recurrent neural network (RNN) parameterized by $\Omega$. That is completely different from some approaches in the literature that have proposed the adoption of recurrent units at the final layers of the target network, such as~\citet{DBLP:journals/corr/HausknechtS15} and \citet{moreno2019performing}. Here, an RNN is adopted not only to predict values that are used to calculate target values during training but also to built a model that is explored to generate the compact structure of ${\cal{RTM}}$ representing previous experiences. The transitions and Q-values stored in $\cal{TM}$ are processed and used to train an RNN, which seeks to learn the similar transitions sets dynamics, and which in turn is applied for both compacting the set of experiences and producing update values to calculate new targets. Figure~\ref{fig:methods_flows-COMPER} illustrates how \methodACRON\ components interact with each other.


In summary, \methodACRON\ uses a CNN to produce Q-value estimates (similar to DQN), but an RNN makes the predictions to obtain target values during the training. Moreover, the training sampling happens over a compact structure representing all previous experiences. At regular intervals, the previous experiences are used to train that RNN, which in turn is also explored to generate a new compact version of the previous experiences set. In the following, we present details about these processes and how we model sets of similar transitions.

\subsection{Modeling the Transition Memory}\label{sec:transitions_memory}


When an agent is training, it experiences successive state transitions through its interactions with the environment, along with respective rewards, while the agent's model is adjusted based on its value function. Therefore, these transitions can be stored in a transition memory (${\cal{TM}}$) to compose a history of expected rewards adjustments associated with each transition (analogously to an experience replay buffer). More specifically, a transitions memory can store the tuples that define the transitions resulting from applying a given policy to each iteration of the training episodes, associated with the long-term rewards estimates for that iterations. Generally, in approaches using experience replay, the transitions are stored in a single set and sampled based on some probability distribution or weight assignment. However, this set brings with it a temporal ordering arising from the insertions of these transitions throughout the iterations. That temporal relationship can be expressed through the evolution of results related to the application of a given policy.


Considering that each episode starts in the initial game state and finishes when the game is over at a final state and the game restart and another episode are carried out, and so on throughout an agent training trial, state transitions can occur more than once over subsequent episodes. So, we can store these transitions in ${\cal{TM}}$. Moreover, we can catch the transitions sample on training and store it in ${\cal{TM}}$ along with its new Q-value estimates (computed on the update step). Thus, probably several transitions in ${\cal{TM}}$ may  be similar in their components (previous state, selected action, immediate reward, and resulting state) and will vary only in their estimates of the value function. The criterion (presented ahead) adopted to consider transitions as similar ones is decisive for modeling $\cal{TM}$. But its application allows us to organize $\cal{TM}$ to identify similar transitions sets and analyze the evolution of their Q-value estimates, allowing \methodACRON\ to predict the Q-values associated with those sets in the next iterations through its RNN. Thus, it can exploit this to produce a more effective fit of the agent model parameters.

\subsubsection{Similar Transitions Sets}\label{sec:mt_as_similar_t}

At each training time-step $t$, we define a transition by a tuple $\tau_t = (s_t, a_t, r_t, s_{t+1})$, in which $s_t$ is the current state, $a_t$ is the action taken at that state, $r_t$ is the received reward at $t$, and $s_{t+1}$ is the resulting state after taking action $a_t$. Consider two distinct transitions $\tau_{t_1} = (s_{t_1}, a_{t_1}, r_{t_1}, s_{{t_1}+1})$ and $\tau_{t_2} = (s_{t_2}, a_{t_2}, r_{t_2}, s_{{t_2}+1}), t_1 \neq t_2$. 
Those two transitions are said to be similar, and 
in that case, we write $\tau_{t_1} \approx \tau_{t_2}$ when the distance between $\tau_{t_1}$ and $\tau_{t_2}$ is lesser than a threshold, i.e., ${\cal{D}}(\tau_{t_1}, \tau_{t_2}) \leq \delta$, in which ${\cal{D}}$ could be any distance measure, e.g., Euclidean distance, and $\delta$ is a distance (or similarity) threshold value.


Let be $N$ the total number of transitions that occurred up to a time instant. Thus, those $N$ transitions are stored in ${\cal{TM}}$ and can be identified as subsets of similar transitions ${\cal{ST}}$ when the similarity condition is satisfied. Moreover, they are stored throughout subsequent agent training episodes and identified by a unique index given transition and all transitions similar to it.

Therefore, we can define ${\cal{TM}}=\left\{[T^i, {\cal{ST}}_i]\,|\,i=1,2,3,\ldots, N_{ST} \right\}$, in which $N_{ST}$ is the total number of distinct subsets of similar transitions, $T^i$ is a unique numbered index and ${\cal{ST}}_i$ represents a set of similar transitions and their Q-values. Thus, 

\begin{equation}
{\cal{ST}}_i = \left\{\left[\tau_{i(1)}, Q_{_i(k)}\right]\;|\; 1 \leq k \leq N^i_{ST} \label{eq:timeseries} \right\} 
\end{equation}

\noindent in which $N^i_{ST}$ represents the total number of similar transitions. Thus, $\tau_{i(1)}$ corresponds to some transition $t_j, j \in \{1, \ldots\ N\}$ and is the first representing transition of similar transitions set ${\cal{ST}}_i$, and $Q_i(k)$ is the Q-values corresponding to some transition $t_j, j \in \{1, \ldots\ N^i_{ST}\}$ such that:

\begin{itemize}
    \item $\tau_{i(1)} \in {\cal{ST}}_i$, and
    \item $\tau_{i(1)} \approx \tau_{i(k_1)} \approx \tau_{i(k_2)}, 1 \leq k_1 < k_2 \leq N^i_{ST}$.
\end{itemize} 
\noindent 

Therefore, ${\cal{TM}}$ can seem as a set of ${\cal{ST}}$. A single representative transition for each ${\cal{ST}}$ can be generated together with the prediction of their next Q-value from an explicit model of ${\cal{ST}}$ using an RNN. In the following, this process is described.

\subsection{Reduced Transition Memory}\label{sec:mt_reduction}



From $\cal{TM}$ we can produce a smaller memory named Reduced Transitions Memory (${\cal{RTM}}$) so that ${\cal{RTM}} = \left\{[\tau'_i]\,|\,i=1,2,3,\ldots, N_{ST} \right\}$, in which $\tau'_i$ is the transition that represents all the similar transitions so far identified in ${\cal{ST}}_i$. Unlike ${\cal{TM}}$, ${\cal{RTM}}$ does not take care about sets of similar transitions, since each $\tau'_i$ is unique and represents all the transitions in a given ${\cal{ST}}_i$. It gives the transitions stored in ${\cal{RTM}}$ the chance of having their Q-values re-estimated. Besides, sampling from ${\cal{RTM}}$ increases the chances of selecting rare and very informative transitions more frequently, at the same time that helps increasing diversity (because of variability in each sample).

One can notice that the similar transitions in ${\cal{TM}}$ correspond to the classic form of a dynamic and recurring system.  Since ${\cal{TM}}$ stores transitions and respective Q-values with a temporal ordering,  we can use that to train a recurrent neural network (RNN) such as a long short-term memory (LSTM)~\citep{doi:10.1162/neco.1997.9.8.1735} by sampling batches of ${\cal{ST}}$. Therefore, we can use that LSTM (that we named QLSTM) to predict the Q-value of the target Q-function. At this point, the ${\cal{RTM}}$ is produced by inserting in this the transitions $\tau'_i$ that represents each ${\cal{ST}}_i$ sampled from ${\cal{TM}}$ to train the QLSTM.

\subsection{Training Q-value LSTM -- QLTSM}\label{sec:QRNN}


In \methodACRON\, the QLSTM has two main functions: (i) learning the transitions history along with their updated Q-values; (ii) predict the next Q-value for a transition in the next time is observed. This attempt to two important processes: (i) reduce the transitions memory and (ii) predict the target Q-value. The process to obtain the similar transitions sets ${\cal{ST}}$ from ${\cal{TM}}$ for training the QLSTM is presented in Equation~\ref{eq:train_qlstm}.

The ${\cal{ST}}$ must be prepared in a way to compose a sequence of transitions and Q-values, trying to describe the policies in their transitions and the evolution of their estimates for the value function. To achieve that new pairs of transition and Q-values are generated aligning each transition with the Q-value estimate that belongs to the following transition, considering the temporal order-preserving mapping. This way, the new pairs of transition and Q-values are easily defined, and a whole training set to use in QLSTM could be defined as follows:\vspace*{-8pt}

\begin{equation}
\mbox{QLSTM--TrainSet} = \bigcup_{i=1}^{N_{ST}} \left\{ [\tau_{o_i(k)}, Q_{o_i(k+1)}]\;|\; 1 \leq k \leq N^i_{ST}-1 \right\}.
\label{eq:train_qlstm}
\end{equation}

The rest of the training procedure of QLSTM is fairly standard and similar to any other RNN, using Mean Square Error (MSE) loss and backpropagation to update the parameters $\Omega$ of the recurrent model. After that, TM is emptied, since QLSTM has incorporated in its model all previous experiences.



\subsection{Identifying and Indexing Similar Transitions}\label{sec:indexing_similar_transitions}

We identify similar transitions by defining a Transitions Memory Index ${\cal{TMI}}$ as part of ${\cal{TM}}$. At each training time-step $t$ we insert a transition ${\tau_{t}}$ together with its estimated Q-value ${Q_t}$ into ${\cal{TM}}$ and check the ${\cal{TMI}}$ for a similar transition given a minimum distance (or similarity) threshold $\delta$. If there is a similar transition, ${\cal{TMI}}$ will return its unique numerical identifier $T^i$. Otherwise, we insert $\tau_{t}$ together with $Q_{t}$ into ${\cal{TMI}}$ and get the corresponding $T^i$. So, a given $T^i$ is used to find the first $\tau_{i(1)}$ similar transition in ${\cal{TM}}$ and update their $Q_{i(1)}$, or to insert $\tau_{t}$ as the first similar transition, together with its first $Q_{t}$, identified by $T^i$. The set of subsequent transitions identified as similar under the same identifier $T^i$ are what we call ${\cal{ST}}_i$. But we don't need to store the entire ${\cal{ST}}_i$. We can use its corresponding $T^i$ to locate its representing transition tuple and update its associated Q-value, considering that its previous value contributions were learned by the QLSTM. Algorithm~\ref{algo:tm-adding-tansition} shows details of that procedure.

\begin{algorithm}[ht]
\SetKwInOut{Input}{input}
\Input{transition $[\tau_{t},Q_t]$, distance (or similarity) threshold value  $\delta$}

${T^i}\leftarrow$ ${\cal{TMI}}$.GetIndex($\tau_{t},\delta$)\;
\SetAlgoLined
 \eIf(){${T^i} == 0$}{\label{lt}
  ${T^i}\leftarrow$ ${\cal{TMI}}$.UpdateIndex($\tau_{t}$)\;
   ${\cal{TM}}$.Insert(${T^i},[\tau_{t},Q_t]$)\;
 }{
   ${\cal{TM}}$.Update(${T^i},Q_t$)\
 }
 \caption{${\cal{TM}}$-StoreTransition}
 \label{algo:tm-adding-tansition}
\end{algorithm}

\subsection{\methodACRON\ -- Detailed Description}\label{sec:theComperMethod}\label{sec:DQNRTM}


\methodACRON\ uses two memories, named Transitions Memory (${\cal{TM}}$) and Reduced Transitions Memory (${\cal{RTM}}$). A convolutional neural network (CNN) approximates the action-value function to estimates the Q-values (\cal{Q\textsubscript{value}}) from a sample of transitions stored in ${\cal{RTM}}$. Another component named QLSTM uses an LSTM network both to produce the ${\cal{RTM}}$ from ${\cal{TM}}$ and to predict the target function value (\cal{Q\textsubscript{target}}), that is inferred from similar transitions sets stored in ${\cal{TM}}$ and used to calculate the temporal-difference error (TDE).

Algorithm~\ref{algo:dqnrtm_algorith} presents the main steps of \methodACRON\ that are carried ou at specific times throughout the iterations of each training episode. The agent observes the current environment state ($s_t$) at each iteration ($t$) and selects an action ($a_t$) through an \textit{$\epsilon$-greedy} strategy. Then, it performs $a_t$ on $s_t$, receives an immediate reward ($r_{t}$), observes a new state ($s_{t+1}$), and an estimate of \cal{Q\textsubscript{value}}. These elements make up the transition $\tau_{t}$ which is indexed by the transitions memory index $({\cal{TMI}})$ and stored in ${\cal{TM}}$. The training of QLSTM and its use to produce ${\cal{RTM}}$ from ${\cal{TM}}$ are performed with the frequency defined by the hyper-parameter $utf$. In turn, the CNN parameters updating is performed at the frequency defined by the hyper-parameter $tf$, and consists of: (i)~performing a uniform sampling of transitions from ${\cal{RTM}}$; (ii)~predicting the \cal{Q\textsubscript{target}} using QLSTM; and (iii)~estimating the \cal{Q\textsubscript{value}} value using the CNN itself. The TDE is calculated from these estimates and used to update the CNN model through backpropagation. 

\begin{algorithm}[ht]
\SetKwInOut{Input}{input}
\Input{batch size $k$, learning rate $\alpha$, trainging frequency $tf$, number of states to be observed $sn$, $\epsilon$-Greedy probability $\epsilon$, update target frequency $utf$, discount factor $\gamma$, similarity threshold value $\delta$}

Initialize $TM  \leftarrow \emptyset$, $RTM \leftarrow \emptyset$, $t \leftarrow 0$, $\Delta\leftarrow0$\;
Initialize $\Theta$, $\Omega$\;
Initialize $env \leftarrow EnvironmentInstance$,  $run \leftarrow True$, $countframes\leftarrow 0$\;
$s_{t} \leftarrow env.InitialState()$\;
 $a_{t}$, $Q_{t}$ $\leftarrow$ \(\epsilon\mbox{-Greedy}\big(s_t,\epsilon, \Theta )\)\;

\SetAlgoLined
 \While{$run$}{
      $s_{t+1},r_t \leftarrow step(s_{t},a_{t})$\;
      ${\cal{TM}}.StoreTransition([\tau(s_{t},a_{t},r_{t},s_{t+1}), Q_t],\delta)$\;
      $s_{t} \leftarrow s_{t+1}$\;
      $t \leftarrow t+1$\;
       \If(){$t$ mod $tf$ == 0}{\label{lt}
       \If(){${\cal{RTM}} == \emptyset$ or $t$ mod $utf$ == 0}{\label{lt}
        $\Omega \leftarrow QLSTM.Train({\cal{TM}})$\;
        ${\cal{RTM}}\leftarrow $QLSTM.ProduceRTM(${\cal{TM}},\Omega$)\;
       }
       \For{$k\leftarrow 1$ \KwTo $K$}{
        Sample $\tau'=(s_{t},a_{t},r_t,s_{t+1}) \sim P(\tau')$ from ${\cal{RTM}}$\;
        $Q\textsubscript{target} \leftarrow QLSTM.QValue(\tau',\Omega)$\;
        $Q\textsubscript{value} \leftarrow  CNN.Forward(s_{t},\Theta)[a_t]$\;
        ${\cal{L}}(\Theta)$ $\leftarrow$ $r_t + \gamma \times Q\textsubscript{target}-Q\textsubscript{value}$\;
        $\Delta \leftarrow \Delta + {\cal{L}}(\Theta) \times \bigtriangledown_\Theta Q\textsubscript{value}$\;
       }
       $\Theta  \leftarrow \Theta + \alpha \times \Delta$\;
       $\Delta\leftarrow 0$\;
       
    }
    $countframes\leftarrow countframes+env.framesnumber$\;
    \If{env.ReachedFinalState()}{
        $run \leftarrow (countframes \leq sn)$\;
        \If{run}{
            $env.ResetStates()$\;
            $s_{t} \leftarrow env.InitialState()$\;
        }
    }
    \(a_{t}\), \(Q_t\)  $\leftarrow$ \(\epsilon\mbox{-Greedy}\big(s_{t},\epsilon, \Theta )\)\;
 }

 \caption{COMPER}
 \label{algo:dqnrtm_algorith}
\end{algorithm}

\begin{algorithm}[h]
\SetKwInOut{Input}{input}
\SetKwInOut{Output}{output}
\Input{State s, $\epsilon$-Greedy probability $\epsilon$, 
CNN parameters $\Theta$}
\Output{action a, Q-value Q}
\SetAlgoLined
 qValues $\leftarrow$ $CNN.Forward(s, \Theta)$\;
 \eIf(){$randonNumber>\epsilon$}{\label{lt}
  a $\leftarrow$ argmax(qValues)\;
  Q $\leftarrow$ qValues[a]\;
 }{
   a $\leftarrow$ random(\#$actions$)\;
   Q $\leftarrow$ qValues[a]\;
 }
 return a, Q\;
 \caption{$\epsilon\mbox{-Greedy}$}
 \label{algo:greedy}
\end{algorithm}

COMPER has the following main parameters as input (see Algorithm~\ref{algo:dqnrtm_algorith}): (i)~$sn$ is the number of states to be observed at each agent training run; (ii)~$k$ is the number of transitions to be sampled from ${\cal{RTM}}$ on line 16 for updating CNN parameters; (iii)~$\alpha$ is the learning rate; (iv)~$\epsilon$ is the probability for selection of actions in Algorithm~\ref{algo:greedy}; (v)~$tf$ represents the frequency with which models will be adjusted -- CNN is always adjusted with that frequency, but QLSTM possibly not (see line 11); (vi)~$utf$ represents the frequency with which the QLSTM is trained and used to update ${\cal{RTM}}$ (see line 12); (vii)~$\gamma$ is a discount factor used in TD error calculation (see line 20); and $\delta$ is a threshold value to define similar transitions. Both memories ${\cal{TM}}$ and ${\cal{RTM}}$ are initialized as empty sets (see line 1). A variable $t$ is used to counts the number of the agent steps. A vector $\Delta$ is used to accumulate the values of the gradient vectors of the loss function (see line 21). Thus, it can be used to adjust CNN parameters $\Theta$ using the learning rate $\alpha$  (see line 23). 

The interaction with the environment starts when an initial state $s_t$ is observed, an action $a_t$ is selected and a $Q_t$ value is estimated, for initial $t==0$ (see lines 4 and 5). From that, the agent's interactions are performed for each $t$ iteration (specifically in lines 7, 9, and 34) until $countframes > sn$ and the last training episode is over (see lines from 26 to 33). A transition occurs when an action $a_t$ is selected from the current state $s_t$ and is taken on the environment, resulting in a reward $r_t$ and the next state $s_{t+1}$ (see line 7). The resulting state $s_{t+1}$ becomes the current state $s_t$ (see line 9), and an action $a_t$ is selected from that state to be taken on the environment (see line 34) at the next interaction. All transitions tuples together with their estimates of Q-values are stored in ${\cal{TM}}$ (see line 8). 

The agent learning process is from lines 7 to 25. The model to estimate $Q_{value}$ is updated from lines 16 to 24, and is carried out at a frequency rate defined by the hyperparameter $tf$. If ${\cal{RTM}}$ is empty or depending on the hyperparameter $utf$, the QLSTM model is updated (lines 12 to 15), that is, its LSTM network is trained (see line 13), and then it is used to update ${\cal{RTM}}$ (see line 14). 

To update the $Q_{value}$ function, a batch of transitions (with the size $k$) is sampled from ${\cal{RTM}}$ (see line 17) and is given as input to the QLSTM, which performs the prediction of $Q_{target}$ for each transition in the sample (see line 18). Then, an estimate of $Q_{value}$ is made using the CNN (see line 19) for the state $s_t$ with action $a_t$ (for each sampled transition). The TDE is calculated from $Q_{value}$ and $Q_{target}$ (see line 20). A gradient vector as a function of TDE is then obtained and accumulated in a vector $\Delta$ (see line 21). After iterating over all the sampled transitions (which is always done in parallel), CNN parameters $\Theta$ are updated at a learning rate $\alpha$ (see line 23). 

Finally, the trade-off between exploration and exploitation to choice of actions is dealt with in Algorithm~\ref{algo:greedy}. Given the current state of the environment, we choose an action from a uniform distribution over the set of possible actions with a given probability. Otherwise, we select the action that maximizes the $Q_{value}$ (approximated by CNN using the current parameters). Detailed information on hyper-parameter values, along with descriptions of the deep neural network architectures used in the experiments, can be found in the appendices.
\section{Methodology of Experiments} \label{sec:metodologia}

Our main objective is to propose a new data-efficient model-free reinforcement learning algorithm to reduce the number of experiences needed during its training concerning the total accumulated reward in the long run. We evaluated the \methodACRON\ agent on the Arcade Learning Environment (ALE)~\footnote{Available at https://github.com/mgbellemare/Arcade-Learning-Environment.}, which was proposed by \citet{bellemare13arcade} and revisited by \citet{machado18arcade}. ALE consists of a development platform and a set of challenges for agents with different reinforcement learning method, which involve problems such as non-determinism, stochasticity, and environments with different values and ranges for awarding rewards. We present a more detailed discussion about this in Apendix~\ref{sec_apendix:ale_challenges_important_aspects}.

According to \citet{machado18arcade}, relevant works in the literature have used different methodologies for the development and evaluation of RL agents, making it difficult to analyze and compare their results. Therefore, the authors proposed a very well-defined methodology for experiments and agent evaluation using ALE, and presented a benchmarking with the experimental results for DQN \citep{mnih2015humanlevel} and SARSA(\(\lambda\))\(+\) blob-PROST~\citep{aamas2016Liang}. Mostly, we follows their propositions for carrying out experiments and agent evaluation. Among these, we highlight the following as a premise for our experiments (see detailed discussion in Appendix~\ref{sec_apendix:ale_challenges_important_aspects}):

\begin{itemize}

    \item Agents should be evaluated over the training data and not in \say{test mode}, because we are interesting to evaluate their learning processes, not their gaming performance.
    
    \item The agent's performance should be evaluated on different check points of the training considering the last episode scores.
    
    \item Agents interact with the game environment in an episodic way. An episode begins by putting the environment to its initial state and ends in a natural final state for the game. 

    \item The main measure of the agent's effectiveness consists of the undiscounted sum of the rewards for each episode.
    
    \item Non-determinism and stochasticity are fundamental issues of reinforcement learning and are explored in ALE using two techniques called \textit{stick actions} and \textit{frame skipping}. We use both with the same configuration as \citet{machado18arcade}-- see Appendix~\ref{sec_apendix:experimental_setup}.
    
    \item Three approaches are commonly used in the literature to the representation of the states of the environment: (i) \textit{color averaging}, (ii) \textit{frame pooling}, and (iii) \textit{frame stacking}. ALE implements color averaging. Frame stacking is not implemented directly by ALE and is an algorithm decision. Color averaging and frame pooling may end up removing the most interesting form of partial observation in ALE, and frame stacking  reduces the degree of partial observability~\citep{machado18arcade}. We use color averaging and make experiments using single frames (that is, not staked) and staked frames.
    
    \item There are differences in the literature on how to summarize and measure the results and on how the agent performance is evaluated based on these results. These questions lead to an important methodological decision, which refers to the moment when each training episode must end. We consider only the signal of \say{game over} as the end of an episode, as strongly recommended by \citet{machado18arcade}.
    
    \item The way in which results are measured and summarized can hamper or improve the analysis of an agent learning process. \citet{machado18arcade} recommended that the result analysis be carried out on a fixed number of the last episodes at different times during an agent training, allowing its learning evolution to be evaluated. The authors also pointed out that the total number of 200 million frames in agents' training has been adopted in the literature, which implies a great computational cost and a lot of time in experiment execution. Therefore, their methodology allows any researcher to report results earlier during experiments, minimizing the problem of agents' evaluation associated with the computational cost.
    
    \item Training an agent over a given number of episodes before evaluating it can yield misleading results, as the duration of each episode can vary widely between different games and as a function of the agent's performance. A more interesting approach is to measure the amount of training data in terms of the total number of environment states (video frames on Atari 2600) observed by an agent, which makes reproducing experiments and comparing the results easier. In that sense, our training trials stop condition is:  if the total frames count equals the total number of frames to observes in a trial, and the game is over (that is, reached a final state), then we stop the run.
    
    \item Interrupting an ongoing episode as soon as the total number of frames is reached, or waiting for the end of the episode does not represent a clear choice in the literature. We ends the training based on a total number of frames and without interrupting the episode until the game over signal as recommended by \citet{machado18arcade}.
    
    \item To allow the assessment of the agents' generalization capacity, only a smaller subset of games should be used to selection and adjustment of hyperparameters, while an agent should be evaluated through their training using the defined values and without further adjustments over the entire set of games~\citep{machado18arcade}.
\end{itemize}

Relevant works in the literature followed evaluation methods different from those proposed by \citet{machado18arcade} or used an earlier version of ALE~\citep{bellemare13arcade}, which has differences in the treatment of stochasticity and non-determinism.~\citet{kaiser2020modelbased} used the 2018 version of ALE and evaluated their agent on a considerably smaller number of frames (409,600) but still using different methodological procedure for agent evaluation. The authors does not clarify if they used the mean of all the episode scores in each training trial or only the last episodes in different checkpoints -- as recommended by \citet{machado18arcade}, not even if they used the stochasticity features of ALE. Besides, the authors used the iterations of the algorithm as a reference to stopping the agent training (not the total number of video frames) and focused their agent evaluation on comparing the number of iterations required to achieve best results. In addition, they have worked on a model-based reinforcement learning method. Therefore, in Section~\ref{results} we present our experimental results in different levels of detail and compare them with the results of our DQN -- developed according to \citet{mnih2015humanlevel}, and with the benchmark presented by \citet{machado18arcade}.

\subsection{Evaluation Procedures of \methodACRON}\label{sec:experimental_procedures} 
\citet{machado18arcade} defined a limit of 200 million frames for each run of agent training. Four checkpoints were defined for the intervals of 10, 50, 100, and 200 million frames. The rewards average and standard deviation were computed over the last 100 episodes preceding each checkpoint. In this way, the authors performed 5 training runs of the DQN on 5 games (Asterix, Beam Rider, Freeway, Seaquest, and Space Invaders) to define parameters, which they used for agent training on all the 60 games available.

We defined a limit of 100 thousand frames for each run. We log the results and checkpoints at every 100 iterations and at the end of the episodes. We divided the number of episodes obtained in each run by 3 to split the results into tertiles. To summarize and evaluate our results, we calculated the average and standard deviation to the scores (i.e., rewards) obtained at the last 5 episodes preceding each checkpoint. These adaptations were necessary because of the smaller number of episodes obtained in each execution due to the limit of the total number of frames, which is 0.0005\% of the total used by \citet{machado18arcade}. In this way, we also redefined the decay rate of the hyperparameter $\epsilon$ (that balances the relationship between exploration and exploitation). Thus, while in \citet{machado18arcade} the value of $\epsilon$ starts at 1.0 and drops to 0.01 over 1 million frames, in our experiments that decay occurs in 90 thousand frames. An important difference is found in the update frequency of the approximate functions. In DQN method, that is performed by copying the weights \(\Theta\) from \(Q(s_t, a_t, \Theta) \) to $\Theta'$ used for target prediction by \(\widehat{Q}(s_{t + 1}, a', \Theta')\) every 4 steps of an agent's interaction with the environment~\citep{machado18arcade}. In \methodACRON, the weights $\Omega$ of the recurrent neural network that estimates target values \(R(\tau_t, \Omega) \) are updated through \textit{backpropagation} similar to what is done for \(Q (s_t, a_t, \Theta)\), however from different updating rules. In this way, the update of the \textit{target} function is performed every 100 steps of an agent's interaction with ALE. In turn, the update of \(Q(s_{t}, a_t, \Theta) \) is performed with the same frequency defined by~\citet{machado18arcade}. Moreover, the weight updating in \methodACRON\ starts after the first 100 steps of the agent interaction with the environment, instead of the 50,000 step warm-up used for the DQN in \citet{machado18arcade}. But in the experiments with our DQN agent, we adopt a warm-up of 1,000 steps. Finally, we evaluate COMPER by representing the states of the environment either with a single frame in a matrix with dimensions of $84 \times 84 \times 1$ or with frames stacked in a matrix of $84 \times 84 \times 4$. In both cases, we just used the luminance values for every two frames combined into one by ALE, using its color averaging method. Our DQN agent is defined as in \citet{mnih2015humanlevel}. About the hyperparameters, only the total number of frames, the decay of the $\epsilon$, the results log frequency, the learning starts, and the target function update frequency were changed to accomplish our experiments. All other parameters remained the same as those by~\citet{machado18arcade}.

To define the values of the hyperparameters, both for COMPER\footnote{Source code will be available soon at ...} and our implementation of DQN\footnote{Source code will be available soon at ...}, we have made only three trials on the game Space Invaders for both methods. Then we used the best values to run the experiments over all the selected games. 

To build the transitions memory index ${\cal{TMI}}$ (Section~\ref{sec:indexing_similar_transitions}) we used the class library named Faiss~\citep{JDH17}, which was developed for efficient similarity search and clustering of dense vectors. According to the authors, it searches in sets of vectors of arbitrary size, up to ones that possibly do not fit in RAM. We used the library's class IndexFlatL2 that applies Euclidean distance and defines the distance threshold (maximum distance to consider two vectors as similar) through a simple parameter, that we named $\delta$ in Algorithm~\ref{algo:tm-adding-tansition}. We used strict similarity with the threshold parameter equal to zero to closely verify our hypotheses about the possibility of identifying similar transitions sets.


\section{Evaluation Results}\label{results}

\begin{figure}
     \centering
     \begin{subfigure}[b]{0.49\textwidth}
         \centering
         \includegraphics[width=\textwidth]{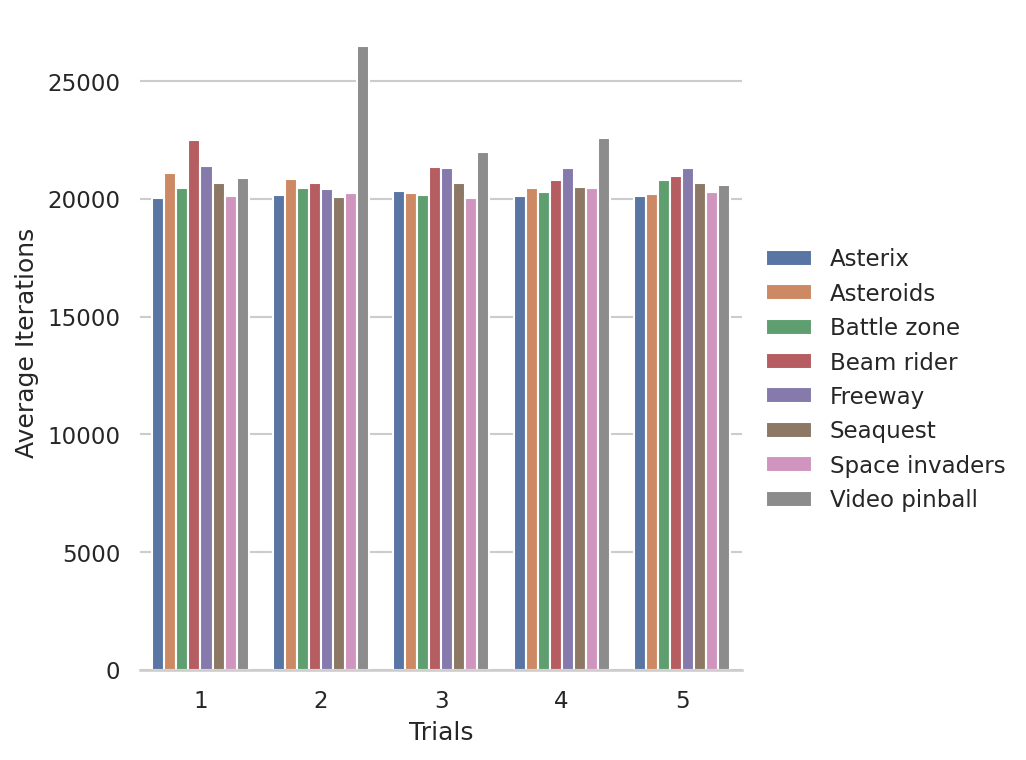}
         \caption{Single Frames - Iterations}
         \label{fig:single_itr_by_trial}
     \end{subfigure}
     \begin{subfigure}[b]{0.49\textwidth}
         \centering
         \includegraphics[width=\textwidth]{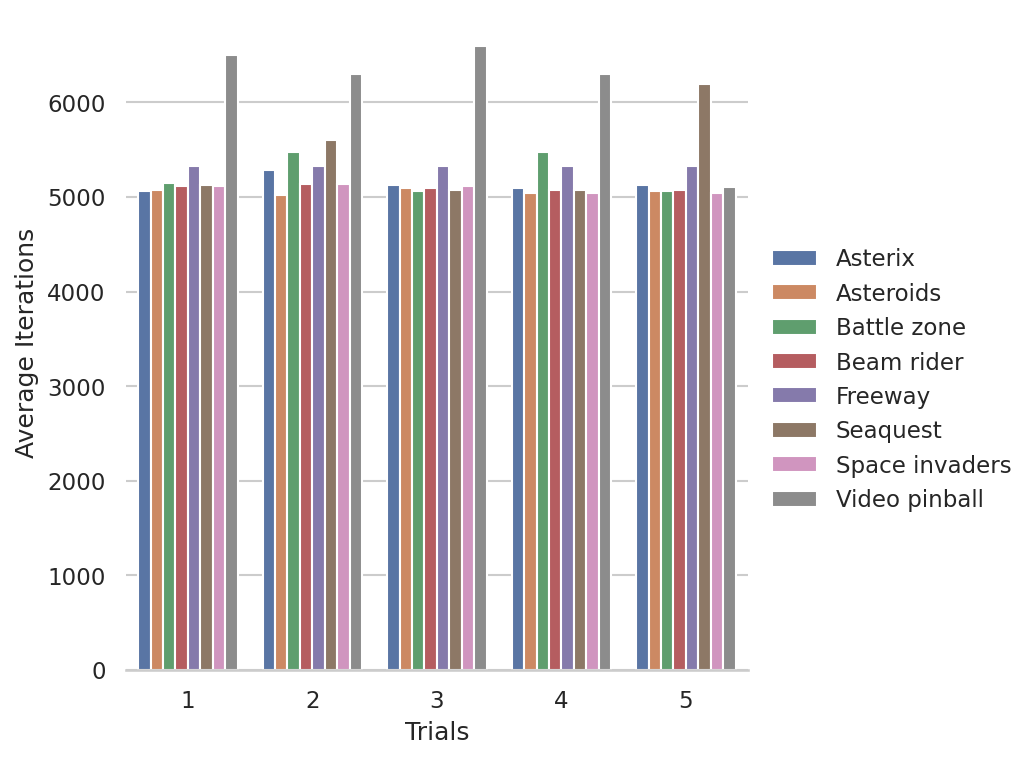}
         \caption{Stacked Frames - Iterations}
         \label{fig:staked_itr_by_trial}
     \end{subfigure}
        \caption{Iterations per Trial}
        \label{fig:iterations_per_trial}
\end{figure}

To evaluate the behavior of \methodACRON\ concerning its learning progress, we report and compare our complete results for all games in five training runs with 100,000 frames and about 25,000 (single frame) and 6,000 (staked frames) iterations, as illustrated in Figure~\ref{fig:iterations_per_trial}. In Section~\ref{subsec:count_similary_transitions}, we analyze the occurrence of similar transitions, while a detailed analysis of the results for each game is presented in Section Section~\ref{subsec:results_sumary}. We used an 8-core CPU (1 thread per core) and 32 GB of RAM to perform our experiments and spent about 9 hours on each training trial using staked frames and 24 hours using single frames.


\subsection{Agent Evaluation Results}\label{subsec:results_sumary}

\begin{table}[b!]
\centering
\begin{tabular}{|c|c|c|c|c|}
\hline
\multirow{2}{*}{Game} & \multicolumn{3}{c|}{\begin{tabular}[c]{@{}c@{}}10\textasciicircum{}5 Frames\\ (10 or 3* last episodes)\end{tabular}} & \begin{tabular}[c]{@{}c@{}}10\textasciicircum{}7 Frames\\ (100 last episodes)\end{tabular} \\ \cline{2-5} 
 & \begin{tabular}[c]{@{}c@{}}COMPER\\ (Single)\end{tabular} & \begin{tabular}[c]{@{}c@{}}COMPER\\ (Stacked)\end{tabular} & DQN & \begin{tabular}[c]{@{}c@{}}DQN\\ Machado et al. (2018)\end{tabular} \\ \hline
Freeway\textsuperscript{*} & \begin{tabular}[c]{@{}c@{}}19.53\\ (9.77)\end{tabular} & \begin{tabular}[c]{@{}c@{}}\textbf{22.33}\\ (0.0)\end{tabular} & \begin{tabular}[c]{@{}c@{}}0.0\\ (0.0)\end{tabular} & \begin{tabular}[c]{@{}c@{}}13.8\\ (8.1)\end{tabular} \\ \hline
Asteroids & \begin{tabular}[c]{@{}c@{}}\textbf{827.8}\\ (142.93)\end{tabular} & \begin{tabular}[c]{@{}c@{}}362.4\\ (52.05)\end{tabular} & \begin{tabular}[c]{@{}c@{}}237.80\\ (47.57)\end{tabular} & \begin{tabular}[c]{@{}c@{}}301.4\\ (14.3)\end{tabular} \\ \hline
Battle Zone\textsuperscript{*} & \begin{tabular}[c]{@{}c@{}}\textbf{4,733.33}\\ (2,080.6)\end{tabular} & \begin{tabular}[c]{@{}c@{}}4,400.0\\ (2,689.59)\end{tabular} &
\begin{tabular}[c]{@{}c@{}}1666.66\\ (1135.29)\end{tabular} &
\begin{tabular}[c]{@{}c@{}}2,428.3\\ (200.4)\end{tabular} \\ \hline
Video Pinball\textsuperscript{*} & \begin{tabular}[c]{@{}c@{}}\textbf{7,353.26}\\ (7,366.18)\end{tabular} & \begin{tabular}[c]{@{}c@{}}2,524.73\\ (84)\end{tabular} & 
\begin{tabular}[c]{@{}c@{}}6936.13\\ (2992.37)\end{tabular}&
\begin{tabular}[c]{@{}c@{}}4,009.3\\ (271.9)\end{tabular} \\ \hline
Seaquest & \begin{tabular}[c]{@{}c@{}}\textbf{279.6}\\ (24.21)\end{tabular} & \begin{tabular}[c]{@{}c@{}}73.2\\ (3.71)\end{tabular} & \begin{tabular}[c]{@{}c@{}}132.00\\ (10.2)\end{tabular} & \begin{tabular}[c]{@{}c@{}}311.5\\ (36.9)\end{tabular} \\ \hline
Asterix & \begin{tabular}[c]{@{}c@{}}239.0\\ (58.3)\end{tabular} & \begin{tabular}[c]{@{}c@{}}\textbf{258.0}\\ (130.64)\end{tabular} & \begin{tabular}[c]{@{}c@{}}182.00\\ (22.05)\end{tabular} & \begin{tabular}[c]{@{}c@{}}1,732.60\\ (314.6)\end{tabular} \\ \hline
Beam Rider\textsuperscript{*} & \begin{tabular}[c]{@{}c@{}}509.06\\ (89.83)\end{tabular} & \begin{tabular}[c]{@{}c@{}}416.53\\  (123.7)\end{tabular} & \begin{tabular}[c]{@{}c@{}}\textbf{563.20}\\ (89.26)\end{tabular} & \begin{tabular}[c]{@{}c@{}}693.9\\ (111.0)\end{tabular} \\ \hline
Space Invaders & \begin{tabular}[c]{@{}c@{}}182.2\\ (68.27)\end{tabular} & \begin{tabular}[c]{@{}c@{}}110.7\\ (53.58)\end{tabular} & \begin{tabular}[c]{@{}c@{}}\textbf{199.80}\\ (49.78)\end{tabular} & \begin{tabular}[c]{@{}c@{}}211.6\\ (14.8)\end{tabular} \\ \hline
\end{tabular}
\caption{Average scores (standard deviation, in parentheses) of the last training episodes. Note that * indicates that last 3 episodes were used to obtain the results. The best results obtained with 100,000 frames are in bold.}
\label{table:results_sumary_comparing}
\end{table}

Table~\ref{table:results_sumary_comparing} presents the average scores (and standard deviation) for five runs of \methodACRON\ -- with single frame and stacked frames, and our implementation of DQN agent, calculated for the last 3 or 10 training episodes with a limit of 100 thousand frames from ALE. For the games Freeway, Battle Zone, Video Pinball, and Beam Rider only the last 3 episodes were used due to a small total number of episodes.~\citet{machado18arcade} have averaged his scores over the last 100 episodes. We defined checkpoints at the end of every tertile on the total frames number throughout the training episodes to verify the agent learning progress. Table~\ref{table:results_tertile_comparing} presents the average scores (and standard deviation) for the last 3 and 5 episodes before these checkpoints, for five training trials with 100,000 frames. 

\begin{table}[t]
\centering
\resizebox{\textwidth}{!}{%
\begin{tabular}{|c|c|c|c|c|c|c|c|c|c|}
\hline
\multirow{2}{*}{Game} & \multicolumn{3}{c|}{\textbf{End of First Tertile}} & \multicolumn{3}{c|}{\textbf{End of Second Tertile}} & \multicolumn{3}{c|}{\textbf{End of Third Tertile}} \\ \cline{2-10} 
 & \begin{tabular}[c]{@{}c@{}}COMPER\\ (Single)\end{tabular} & \begin{tabular}[c]{@{}c@{}}COMPER\\ (Stacked)\end{tabular} & DQN & \begin{tabular}[c]{@{}c@{}}COMPER\\ (Single)\end{tabular} & \begin{tabular}[c]{@{}c@{}}COMPER\\ (Stacked)\end{tabular} & DQN & \begin{tabular}[c]{@{}c@{}}COMPER\\ (Single)\end{tabular} & \begin{tabular}[c]{@{}c@{}}COMPER\\ (Stacked)\end{tabular} & DQN \\ \hline
Freeway\textsuperscript{*} & \textbf{\begin{tabular}[c]{@{}c@{}}4.26\\ (2.38)\end{tabular}} & \begin{tabular}[c]{@{}c@{}}4.0\\ (0.0)\end{tabular} & \begin{tabular}[c]{@{}c@{}}0.0\\ (0.0)\end{tabular} & \begin{tabular}[c]{@{}c@{}}12\\ (6.70)\end{tabular} & \textbf{\begin{tabular}[c]{@{}c@{}}16.0\\ (0.0)\end{tabular}} & \begin{tabular}[c]{@{}c@{}}0.0\\ (00.)\end{tabular} & \begin{tabular}[c]{@{}c@{}}19.53\\ (10.92)\end{tabular} & \textbf{\begin{tabular}[c]{@{}c@{}}22.33\\ (0.0)\end{tabular}} & \begin{tabular}[c]{@{}c@{}}0.0\\ (0.0)\end{tabular} \\ \hline
Asteroids & \textbf{\begin{tabular}[c]{@{}c@{}}885.20\\ (124.95)\end{tabular}} & \begin{tabular}[c]{@{}c@{}}440.8\\ (85.75)\end{tabular} & \begin{tabular}[c]{@{}c@{}}376.00\\ (81.47)\end{tabular} & \textbf{\begin{tabular}[c]{@{}c@{}}902.40\\ (175.34)\end{tabular}} & \begin{tabular}[c]{@{}c@{}}334.80\\ (50.56)\end{tabular} & \begin{tabular}[c]{@{}c@{}}386.00\\ (51.51)\end{tabular} & \textbf{\begin{tabular}[c]{@{}c@{}}846.0\\ (237.37)\end{tabular}} & \begin{tabular}[c]{@{}c@{}}341.20\\ (78.55)\end{tabular} & \begin{tabular}[c]{@{}c@{}}233.60\\ (0.0)\end{tabular} \\ \hline
Battle Zone\textsuperscript{*} & \begin{tabular}[c]{@{}c@{}}1933.33\\ (862.81)\end{tabular} & \textbf{\begin{tabular}[c]{@{}c@{}}3400.00\\ (1673.32)\end{tabular}} & \begin{tabular}[c]{@{}c@{}}3200.00\\ (2089.65)\end{tabular} & \textbf{\begin{tabular}[c]{@{}c@{}}4333.33\\ (2392.11)\end{tabular}} & \begin{tabular}[c]{@{}c@{}}2333.33\\ (408.24)\end{tabular} & \begin{tabular}[c]{@{}c@{}}2066.66\\ (1038.16)\end{tabular} & \textbf{\begin{tabular}[c]{@{}c@{}}4733.33\\ (2326.17)\end{tabular}} & \begin{tabular}[c]{@{}c@{}}4400.00\\ (3003.70)\end{tabular} & \begin{tabular}[c]{@{}c@{}}1666.66\\ (1269.29)\end{tabular} \\ \hline
Video Pinball\textsuperscript{*} & \begin{tabular}[c]{@{}c@{}}10157.73\\ (6695.87)\end{tabular} & \begin{tabular}[c]{@{}c@{}}2565.73\\ (1442.02)\end{tabular} & \textbf{\begin{tabular}[c]{@{}c@{}}11513.86\\ (7672.34)\end{tabular}} & \begin{tabular}[c]{@{}c@{}}8747.33\\ (8058.55)\end{tabular} & \begin{tabular}[c]{@{}c@{}}2665.46\\ (2844.67)\end{tabular} & \begin{tabular}[c]{@{}c@{}}4454.66\\ (1127.78)\end{tabular} & \textbf{\begin{tabular}[c]{@{}c@{}}8544.60\\ (10671.12)\end{tabular}} & \textbf{\begin{tabular}[c]{@{}c@{}}2524.73\\ (2764.71)\end{tabular}} & \begin{tabular}[c]{@{}c@{}}3936.13\\ (3345.57)\end{tabular} \\ \hline
Seaquest & \textbf{\begin{tabular}[c]{@{}c@{}}143.2\\ (17.52)\end{tabular}} & \begin{tabular}[c]{@{}c@{}}76.8\\ (15.59)\end{tabular} & \begin{tabular}[c]{@{}c@{}}78.40\\ (40.52)\end{tabular} & \textbf{\begin{tabular}[c]{@{}c@{}}356.0\\ (91.82)\end{tabular}} & \begin{tabular}[c]{@{}c@{}}80.0\\ (22.44)\end{tabular} & \begin{tabular}[c]{@{}c@{}}152.00\\ (23.50)\end{tabular} & \textbf{\begin{tabular}[c]{@{}c@{}}222.4\\ (38.32)\end{tabular}} & \begin{tabular}[c]{@{}c@{}}72.0\\ (4.0)\end{tabular} & \begin{tabular}[c]{@{}c@{}}123.20\\ (15.88)\end{tabular} \\ \hline
Asterix & \begin{tabular}[c]{@{}c@{}}268.0\\ (82.28)\end{tabular} & \begin{tabular}[c]{@{}c@{}}258.0\\ (82.88)\end{tabular} & \textbf{\begin{tabular}[c]{@{}c@{}}290.00\\ (58.99)\end{tabular}} & \begin{tabular}[c]{@{}c@{}}228.0\\ (30.33)\end{tabular} & \begin{tabular}[c]{@{}c@{}}210.0\\ (24.49)\end{tabular} & \textbf{\begin{tabular}[c]{@{}c@{}}234.00\\ (18.55)\end{tabular}} & \begin{tabular}[c]{@{}c@{}}232.0\\ (52.15)\end{tabular} & \textbf{\begin{tabular}[c]{@{}c@{}}256.0.0\\ (165.92)\end{tabular}} & \begin{tabular}[c]{@{}c@{}}166.00\\ (45.87)\end{tabular} \\ \hline
Beam Rider\textsuperscript{*} & \textbf{\begin{tabular}[c]{@{}c@{}}417.33\\ (119.26)\end{tabular}} & \begin{tabular}[c]{@{}c@{}}352.53\\ (102.05)\end{tabular} & \begin{tabular}[c]{@{}c@{}}381.33\\ (46.38)\end{tabular} & \begin{tabular}[c]{@{}c@{}}419.46\\ (49.30)\end{tabular} & \begin{tabular}[c]{@{}c@{}}354.93\\ (116.59)\end{tabular} & \textbf{\begin{tabular}[c]{@{}c@{}}475.20\\ (56.12)\end{tabular}} & \begin{tabular}[c]{@{}c@{}}509.06\\ (100.43)\end{tabular} & \begin{tabular}[c]{@{}c@{}}416.53\\ (137.81)\end{tabular} & \textbf{\begin{tabular}[c]{@{}c@{}}563.20\\ (89.26)\end{tabular}} \\ \hline
Space Invaders & \begin{tabular}[c]{@{}c@{}}139.8\\ (42.24)\end{tabular} & \begin{tabular}[c]{@{}c@{}}109.6\\ (42.84)\end{tabular} & \textbf{\begin{tabular}[c]{@{}c@{}}140.80\\ (59.94)\end{tabular}} & \textbf{\begin{tabular}[c]{@{}c@{}}201.6\\ (30.58)\end{tabular}} & \begin{tabular}[c]{@{}c@{}}128.4\\ (31.54)\end{tabular} & \begin{tabular}[c]{@{}c@{}}174.00\\ (35.57)\end{tabular} & \begin{tabular}[c]{@{}c@{}}193.8\\ (63.08)\end{tabular} & \begin{tabular}[c]{@{}c@{}}129.2\\ (75.74)\end{tabular} & \textbf{\begin{tabular}[c]{@{}c@{}}213.80\\ (68.90)\end{tabular}} \\ \hline

\end{tabular}}
\caption{Average scores (standard deviation, in parentheses) of the last episodes of each tertile. Note that \textsuperscript{*} indicates that last 3 episodes were used to obtain the results. The best results obtained for each tertile are in bold.}
\label{table:results_tertile_comparing}
\end{table}

\methodACRON\ obtains the best results on the games Freeway, Asteroids, Battle Zone, and Video Pinball than the DQN with 100,000 and 10 million frames. We highlight Asteroids that was better even compared with 50, 100, and 200 million frames~\citep{machado18arcade}. The large standard deviation presented in Video Pinball, with single frames, is due to a training run for which the agent has not scored in the last three episodes. In Seaquest, the \methodACRON\ results are better than DQN's for 100,000 frames and statistically equivalent for 10 million. For Asterix, the number of frames strongly influenced the agent's performance, but for 100,00 frames \methodACRON\ achieved better results. We consider the results for Beam Rider to be statistically equivalent and for Space Invaders \methodACRON\ performed the worst. Regarding the evaluation on the checkpoints, we highlight the continuous and stable evolution of \methodACRON\ on the games Freeway, Asteroids, Seaquest, Asterix, and Beam Rider.

Freeway, Seaquest, and Asteroids are complex games. The first two are due to the difficulty in rewarding and long-term policy learning needs, and Asteroids is due to the hard exploration task and the complexity in its dynamics according to~\citet{machado18arcade}. The difference of results for Asterix with 10 million frames seems to be related to the high degree of randomness in the appearance of reward elements, making the rewards accumulation strongly related to the number of interactions with the game environment and lesser dependent on the learning of a long-term action policy. On the other hand, there are a lot of simultaneous rewarding elements in the Space Invaders scene from the first frames until long after the game starts. Thus, the agent has many chances to hit an enemy and be hit, and the number of frames does not seem to have a considerable impact on the total reward accumulated up to the limit of 10 million frames, contributing more towards reducing the standard deviation.

Finally, we highlight that \methodACRON\ scores better when using single frames instead of stacked frames and that the training runs with single frames were the ones that produced more similar transitions (see Section~\ref{subsec:count_similary_transitions}). Moreover, the size of the memory from which the agent sampled transitions is considerably lesser than used in the literature. This indicates that:  (1) our hypotheses are probably correct; (2) the agent can learn from smaller memories; and (3) similar transitions' sets capture dynamics of the transitions regarding the long-term expected rewards. 


\subsection{Similar Transitions}\label{subsec:count_similary_transitions}

To verify the behavior of the Transitions Memory ($\cal{TM}$), we computed the number of transitions identified as similar under the same set throughout the training episodes, and the sizes of these sets, as its transitions are removed from the memory when sampled to train the target network, and inserted into the same set if they occur again. 

\begin{figure}[h]
     \centering
     \begin{subfigure}[b]{0.49\textwidth}
         \centering
         \includegraphics[width=\textwidth]{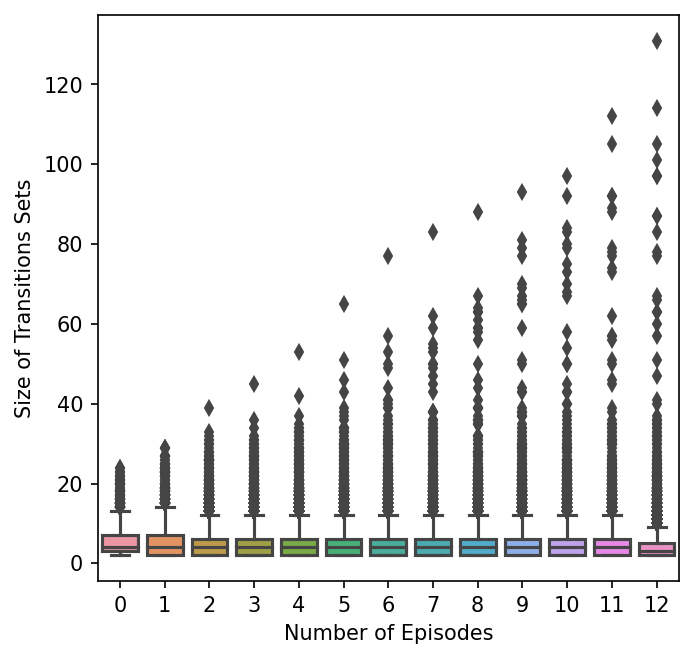}
         \caption{Freeway ($84\times84\times1$)}
         \label{fig:freeway_sizes_simcount_single}
     \end{subfigure}
     \begin{subfigure}[b]{0.49\textwidth}
         \centering
         \includegraphics[width=\textwidth]{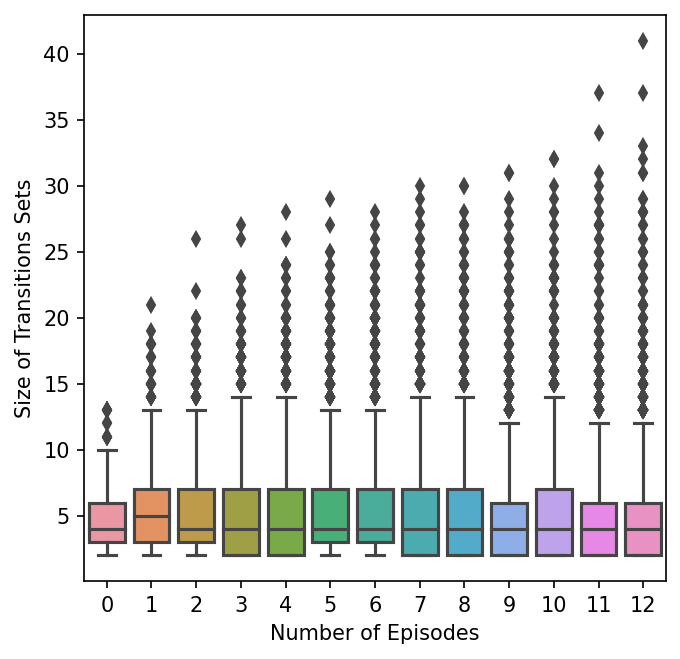}
         \caption{Freeway ($84\times84\times4$)}
         \label{freeway_sizes_simcount__staked}
     \end{subfigure}
     \begin{subfigure}[b]{0.49\textwidth}
         \centering
         \includegraphics[width=\textwidth]{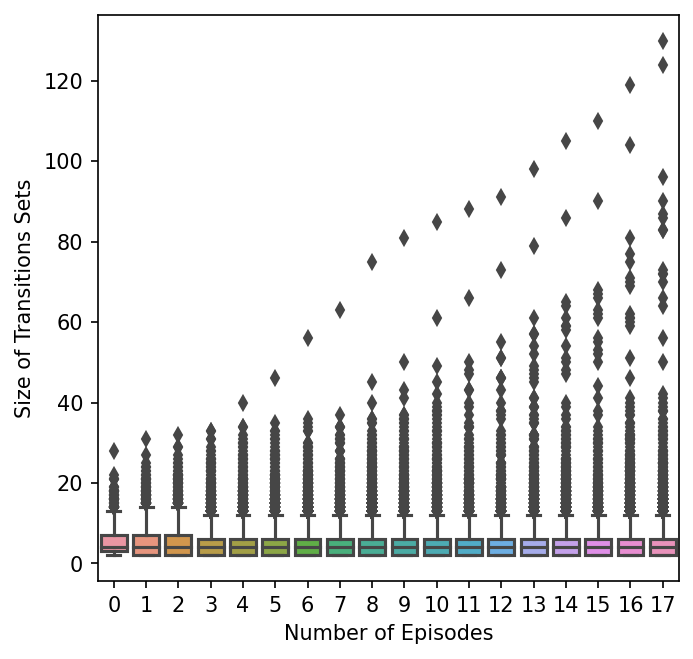}
         \caption{Beam Rider ($84\times84\times1$)}
         \label{fig:beamreider_sizes_simcount_single}
     \end{subfigure}
     \begin{subfigure}[b]{0.49\textwidth}
         \centering
         \includegraphics[width=\textwidth]{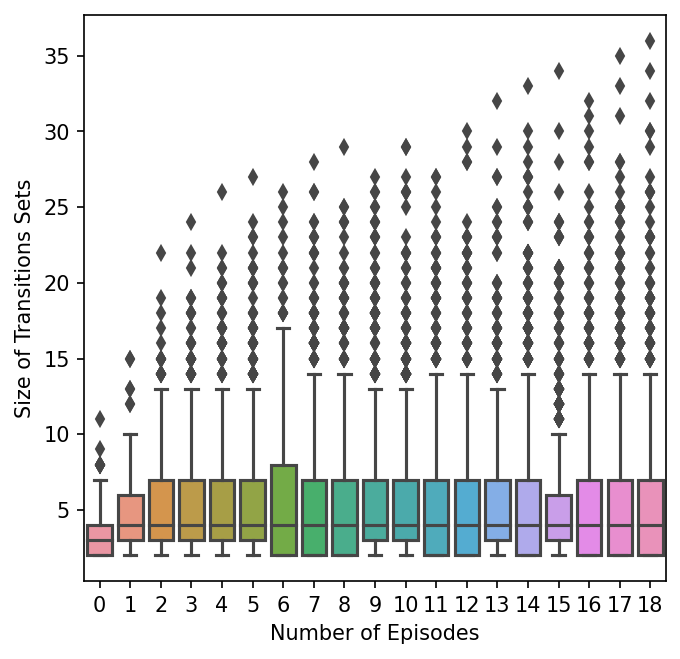}
         \caption{Beam Rider ($84\times84\times4$)}
         \label{fig:beamreider_simcount__staked}
     \end{subfigure}
        \caption{Sizes of Similar Transitions Sets}
        \label{fig:sizes_stes_sim_count}
\end{figure}

Figure~\ref{fig:sizes_stes_sim_count} illustrate the average sizes of the similar transitions sets for Freeway and Beam Rider. We can verify tree main behaviors: (1) new sets were created in different moments; (2) some of these sets are removed from the memory and do not occur again, while others remain being updated until the last training episode; (3) when using single frame, the number of similar transitions is larger than when using staked frames. This behavior is observed in all games.

\begin{figure}[h]
     \centering
     \begin{subfigure}[b]{0.49\textwidth}
         \centering
         \includegraphics[width=\textwidth]{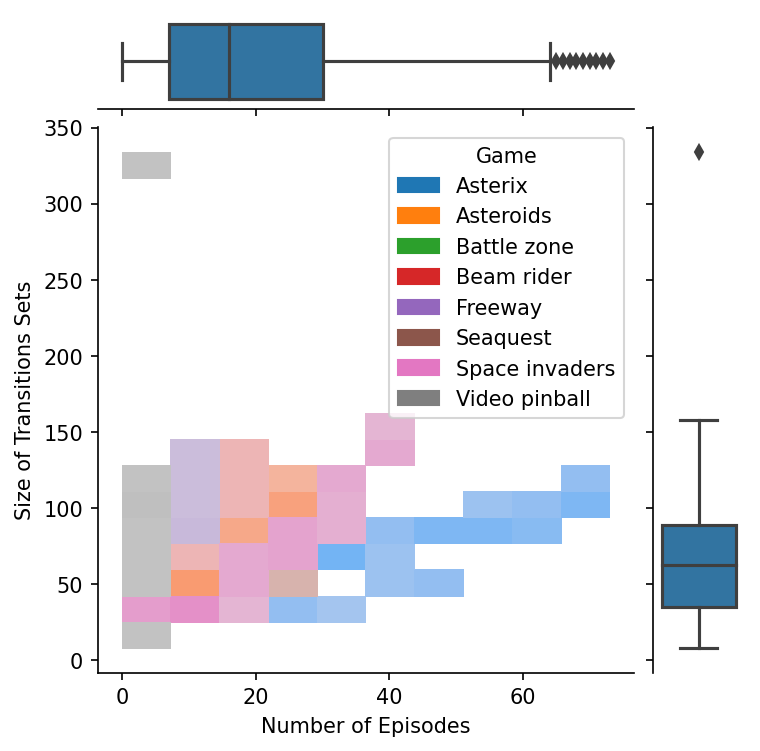}
         \caption{Single Frame ($84\times84\times1$)}
         \label{fig:dist_sizes_simcount_single}
     \end{subfigure}
     \begin{subfigure}[b]{0.49\textwidth}
         \centering
         \includegraphics[width=\textwidth]{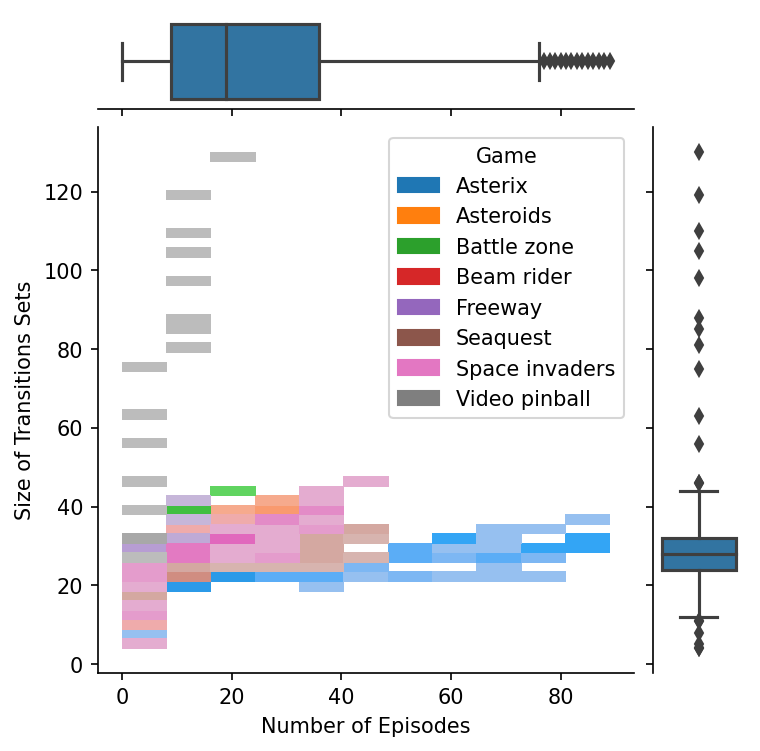}
         \caption{Stacked Frames($84\times84\times4$)}
         \label{fig:dist_sizes_simcount_staked}
     \end{subfigure}
        \caption{Sizes of the Similar Transitions' Sets Throughout Episodes}
        \label{fig:single_staked_sizes_simcount}
\end{figure}

\begin{figure}
     \centering
     \begin{subfigure}[b]{0.49\textwidth}
         \centering
         \includegraphics[width=\textwidth]{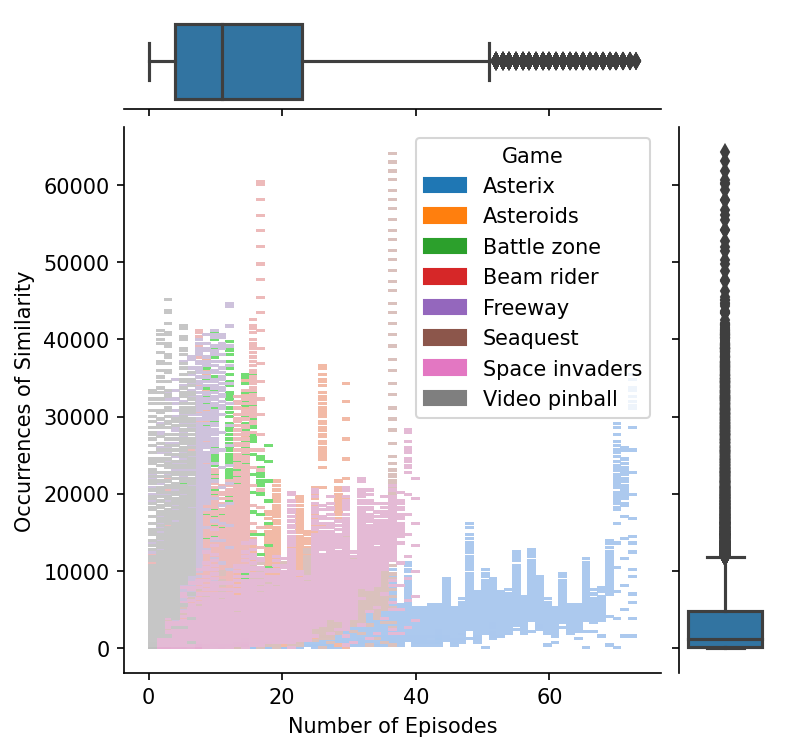}
         \caption{Single Frame ($84\times84\times1$)}
         \label{fig:occurences_of_sim_single_single}
     \end{subfigure}
     \begin{subfigure}[b]{0.49\textwidth}
         \centering
         \includegraphics[width=\textwidth]{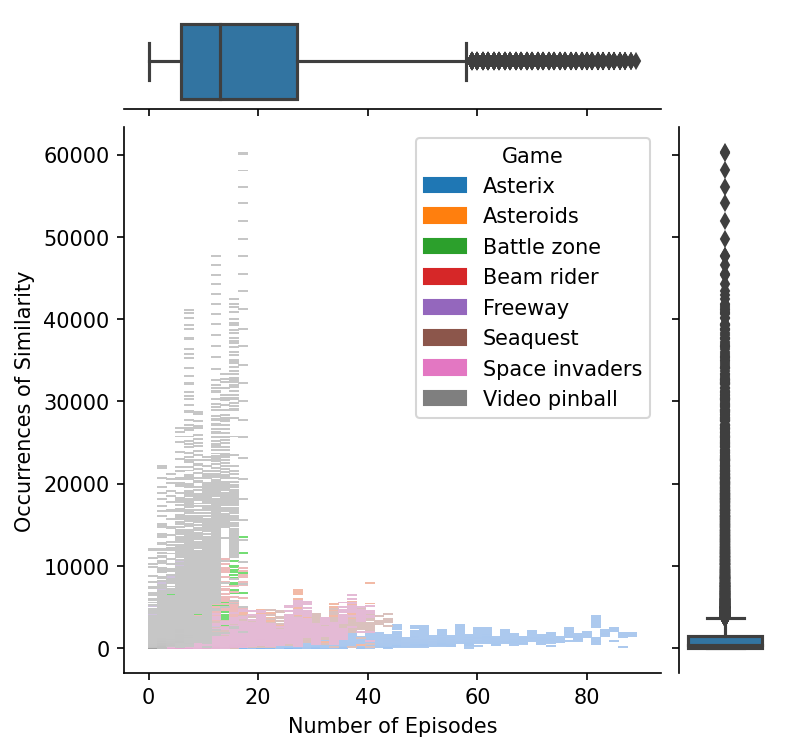}
         \caption{Stacked Frames ($84\times84\times4$)}
         \label{fig:occurences_of_sim_staked}
     \end{subfigure}
        \caption{Occurrences of Similarity}
        \label{fig:occurrences_of_sim__single_staked}
\end{figure}

Figure~\ref{fig:single_staked_sizes_simcount} illustrates the sizes of the similar transitions' sets for all the games with single and stacked frames and Figure~\ref{fig:occurrences_of_sim__single_staked} shows the occurrence of similarities. This information is relevant because the number of similarities implies the number of times the estimate of the Q-Value for a given transition was updated instead of reinserting a new tuple similar to one that already exists in the memory. And that helps to maintain its size small as discussed in Section~\ref{sec:mt_reduction}.

\begin{figure}
     \centering
     \begin{subfigure}[b]{0.49\textwidth}
         \centering
         \includegraphics[width=\textwidth]{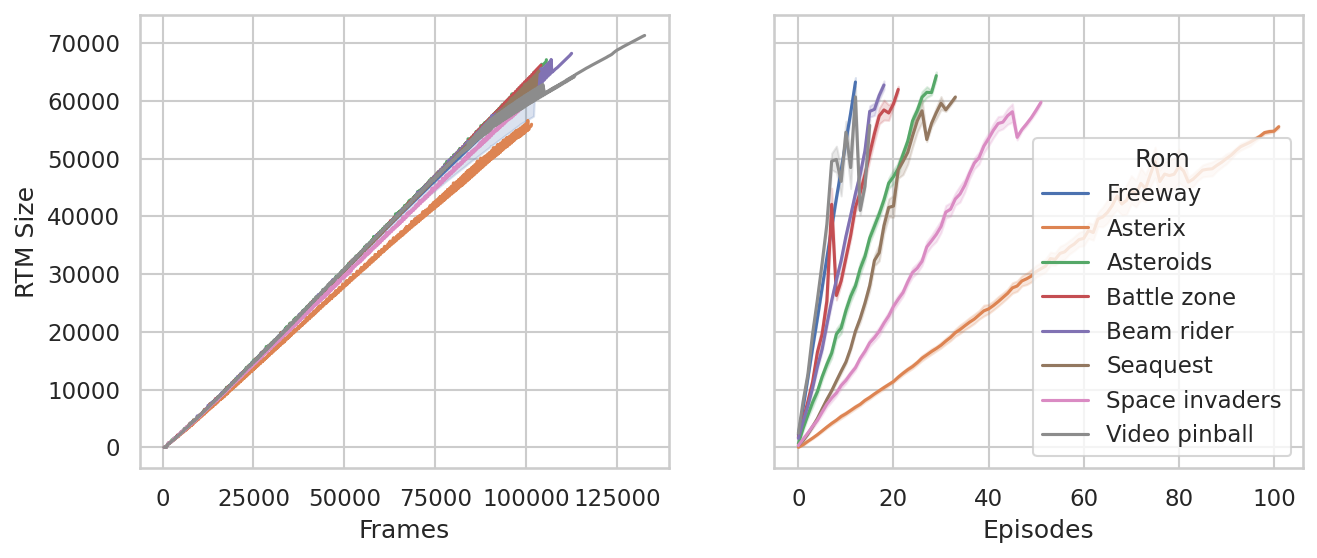}
         \label{fig:single_rtm_size}
     \end{subfigure}
     \begin{subfigure}[b]{0.49\textwidth}
         \centering
         \includegraphics[width=\textwidth]{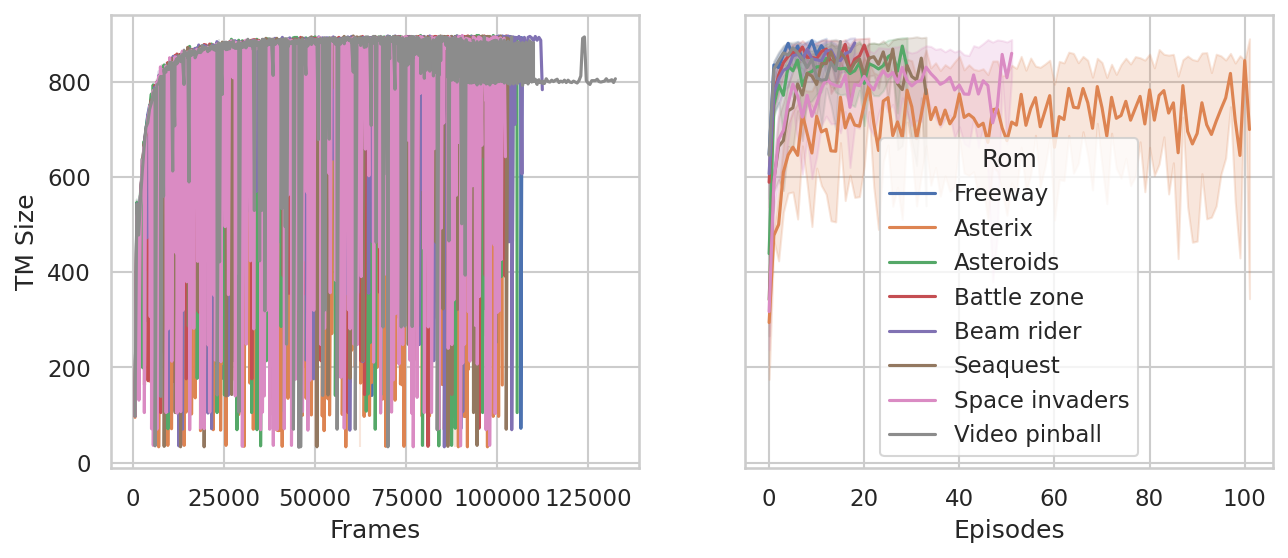}
         \label{fig:single_tm_size}
     \end{subfigure}
        \caption{Behavior of Transition Memories - Single Frame ($84\times84\times1$)}
        \label{fig:single_rtm_tm_sizes}
\end{figure}

\begin{figure}
     \centering
     \begin{subfigure}[b]{0.49\textwidth}
         \centering
         \includegraphics[width=\textwidth]{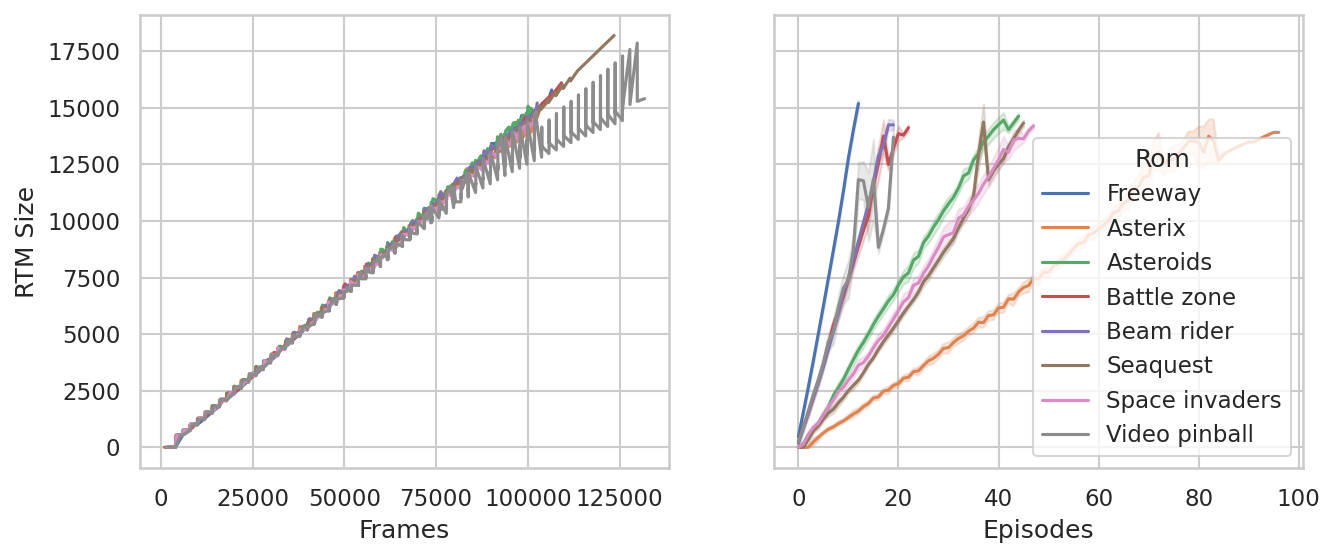}
         \label{fig:staked_rtm_size}
     \end{subfigure}
    \hfill
     \begin{subfigure}[b]{0.49\textwidth}
         \centering
         \includegraphics[width=\textwidth]{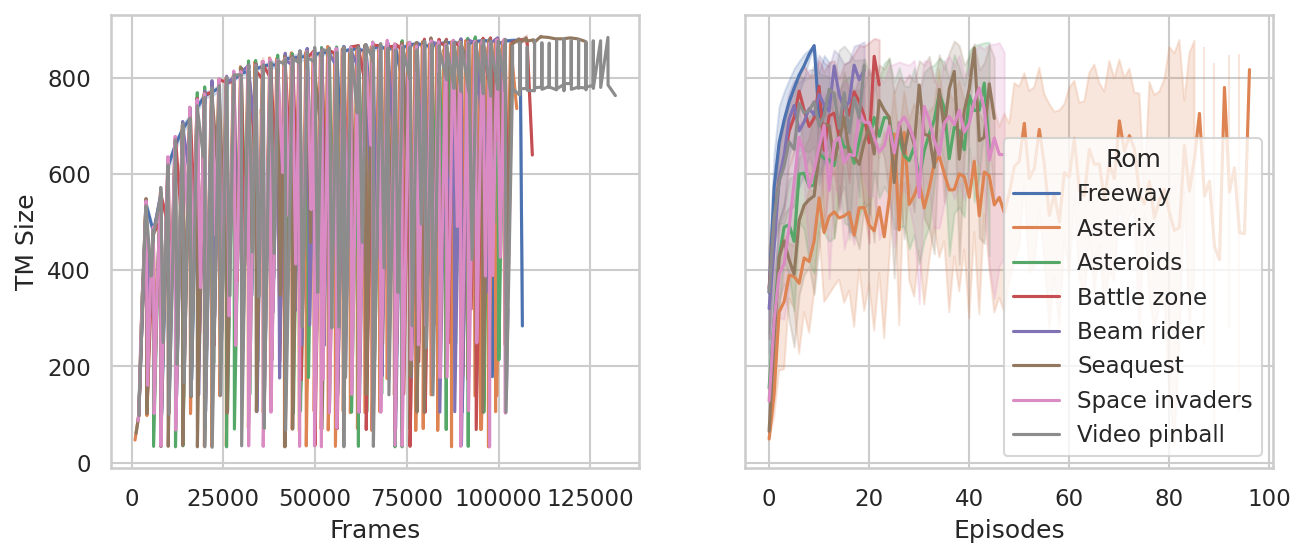}
         \label{fig:stacked_tm_size}
     \end{subfigure}
        \caption{Behavior of Transition Memories - Stacked Frames ($84\times84\times4$)}
        \label{fig:stacked_rtm_tm_sizes}
\end{figure}

As the total number of observed frames increases throughout the successive episodes of each training trial, we can see that the size of $\cal{TM}$ varies but keeps an upper limit because of the reduction step applied when similar transitions are removed to train the LSTM and update $\cal{RTM}$. The small size of  $\cal{RTM}$ is verified directly by the mean of its total number of unique transitions at the end of each trial. Even sampling such a small number of transitions, COMPER achieves good results. This behaviors are ilustrated on Figures~\ref{fig:single_rtm_tm_sizes} and~\ref{fig:stacked_rtm_tm_sizes}.

\subsection{Detailed Results}\label{subsec:results_details}

\begin{figure}[]
     \centering
     \begin{subfigure}[b]{1.0\textwidth}
         \centering
         \includegraphics[width=\textwidth]{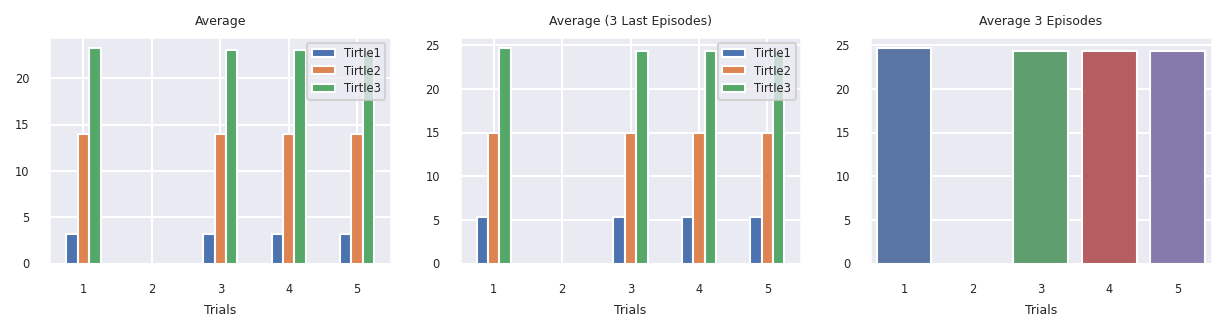}
         \caption{Single Frame}
         \label{fig:results_freeway_single_2}
     \end{subfigure}
     \begin{subfigure}[b]{1.0\textwidth}
         \centering
         \includegraphics[width=\textwidth]{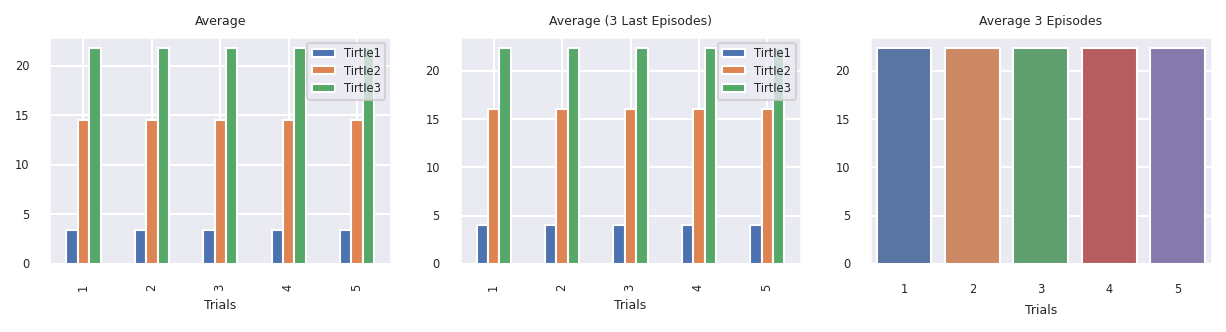}
         \caption{Stacked Frames}
         \label{fig:results_freeway_staked_2}
     \end{subfigure}
        \caption{Detailed results for Freeway}
        \label{fig:results_freeway_detailed}
\end{figure}

\begin{figure}[h]
     \centering
     \begin{subfigure}[b]{1.0\textwidth}
         \centering
         \includegraphics[width=\textwidth]{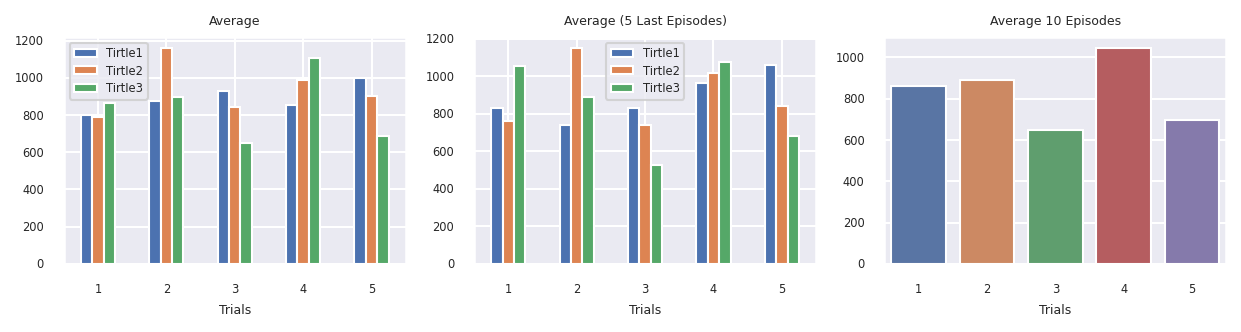}
         \caption{Single Frame}
         \label{fig:results_asteroids_single_2}
     \end{subfigure}
     \begin{subfigure}[b]{1.0\textwidth}
         \centering
         \includegraphics[width=\textwidth]{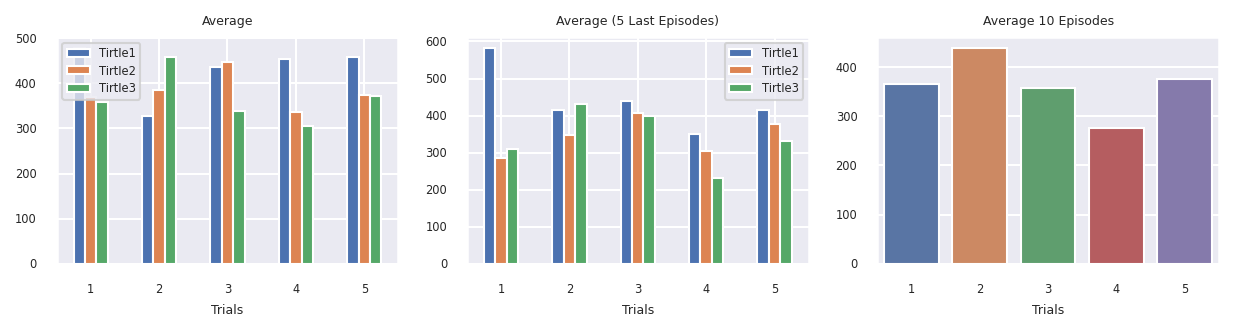}
         \caption{Stacked Frames}
         \label{fig:results_asteroids_staked_2}
     \end{subfigure}
        \caption{Detailed results for Asteroids}
        \label{fig:results_asteroids_detailed}
\end{figure}

\begin{figure}[]
     \centering
     \begin{subfigure}[b]{1.0\textwidth}
         \centering
         \includegraphics[width=\textwidth]{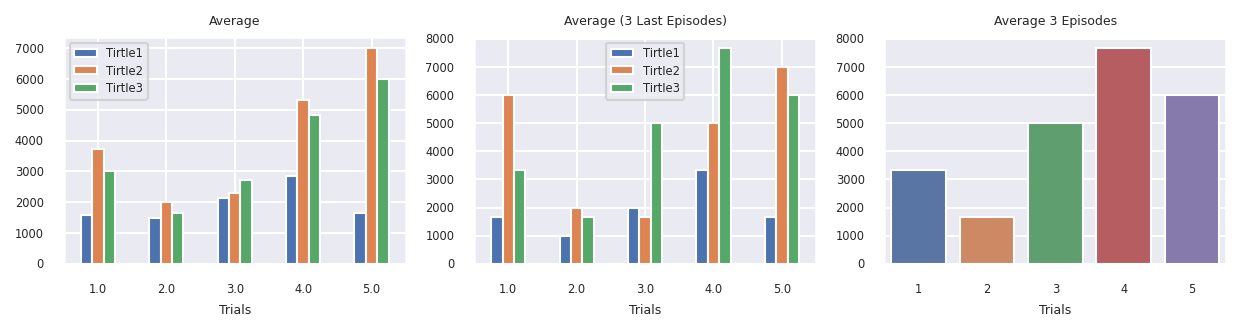}
         \caption{Single Frame}
         \label{fig:results_battle_single_2}
     \end{subfigure}
     \begin{subfigure}[b]{1.0\textwidth}
         \centering
         \includegraphics[width=\textwidth]{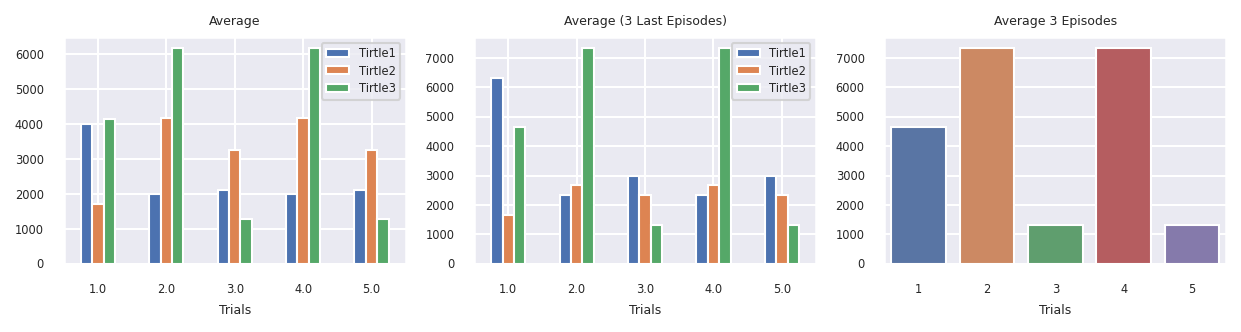}
         \caption{Stacked Frames}
         \label{fig:results_battle_staked_2}
     \end{subfigure}
        \caption{Detailed results for Battle Zone}
        \label{fig:results_battle_detailed}
\end{figure}

\begin{figure}[]
     \centering
     \begin{subfigure}[b]{1.0\textwidth}
         \centering
         \includegraphics[width=\textwidth]{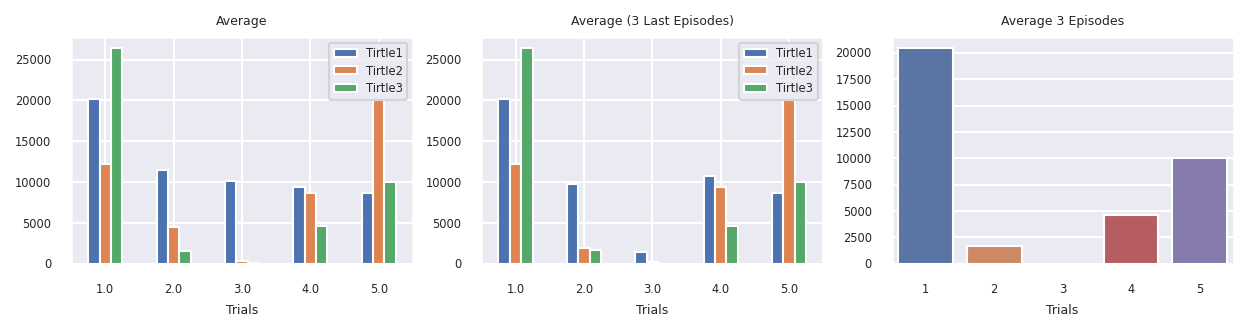}
         \caption{Single Frame}
         \label{fig:results_video_single_2}
     \end{subfigure}
     \begin{subfigure}[b]{1.0\textwidth}
         \centering
         \includegraphics[width=\textwidth]{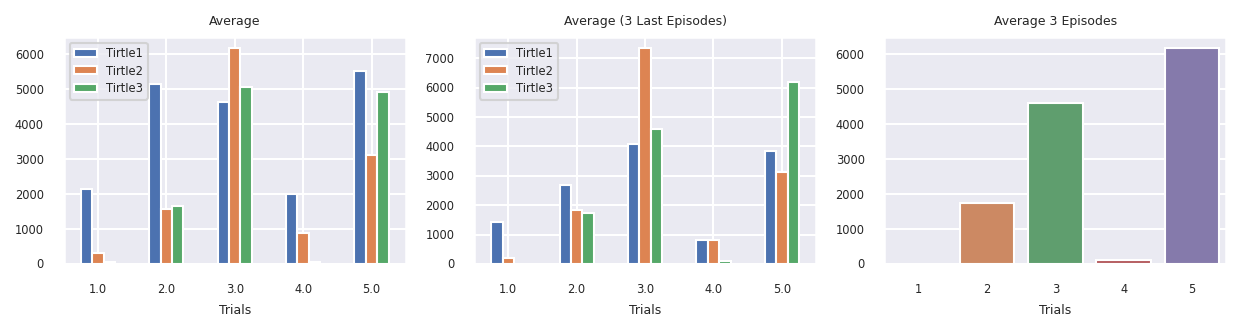}
         \caption{Stacked Frames}
         \label{fig:results_video_staked_2}
     \end{subfigure}
        \caption{Detailed results for Video Pinball}
        \label{fig:results_video_detailed}
\end{figure}

\begin{figure}[]
     \centering
     \begin{subfigure}[b]{1.0\textwidth}
         \centering
         \includegraphics[width=\textwidth]{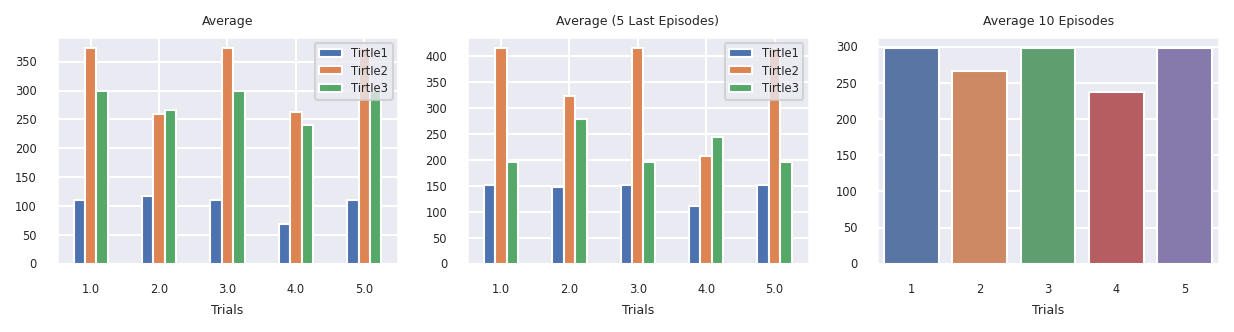}
         \caption{Single Frame}
         \label{fig:results_seaquest_single_2}
     \end{subfigure}
     \begin{subfigure}[b]{1.0\textwidth}
         \centering
         \includegraphics[width=\textwidth]{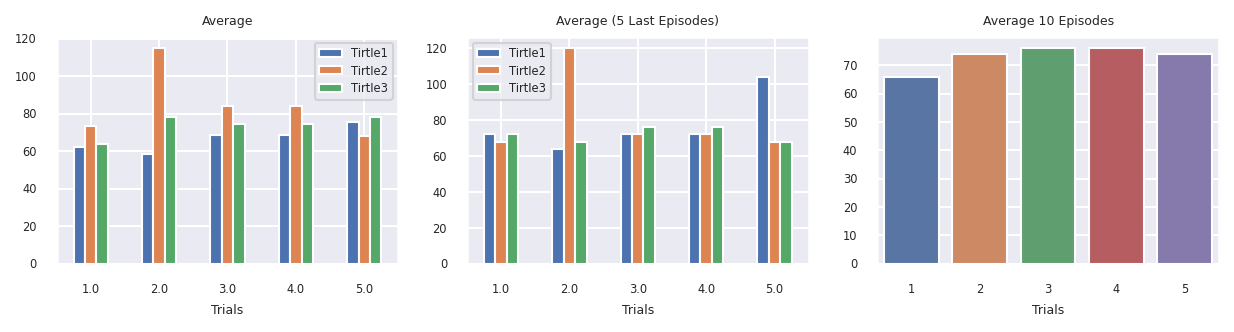}
         \caption{Stacked Frames}
         \label{fig:results_seaquest_staked_2}
     \end{subfigure}
        \caption{Detailed results for Seaquest}
        \label{fig:results_seaquest_detailed}
\end{figure}

\begin{figure}[]
     \centering
     \begin{subfigure}[b]{1.0\textwidth}
         \centering
         \includegraphics[width=\textwidth]{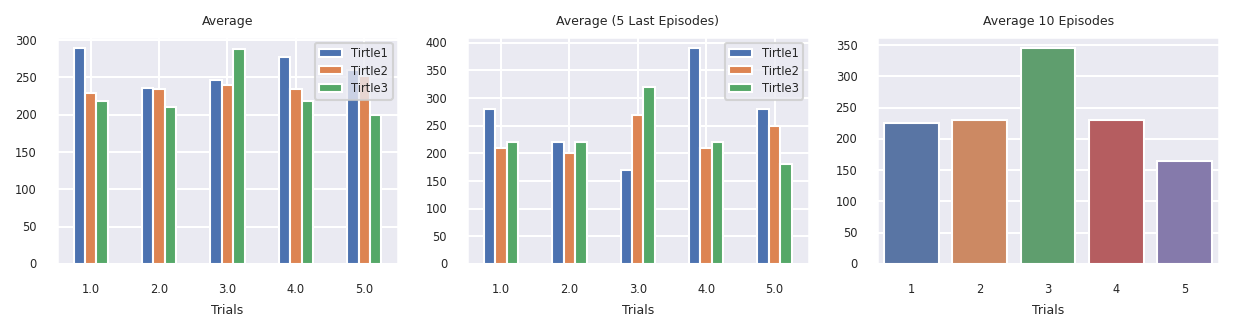}
         \caption{Single Frame}
         \label{fig:results_asterix_single_2}
     \end{subfigure}
     \begin{subfigure}[b]{1.0\textwidth}
         \centering
         \includegraphics[width=\textwidth]{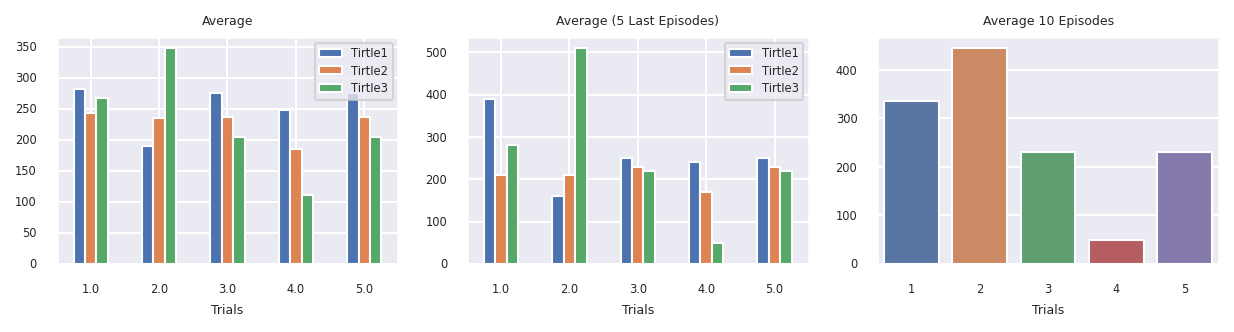}
         \caption{Stacked Frames}
         \label{fig:results_asterix_staked_2}
     \end{subfigure}
        \caption{Detailed results for Asterix}
        \label{fig:results_asterix_detailed}
\end{figure}

\begin{figure}[]
     \centering
     \begin{subfigure}[b]{1.0\textwidth}
         \centering
         \includegraphics[width=\textwidth]{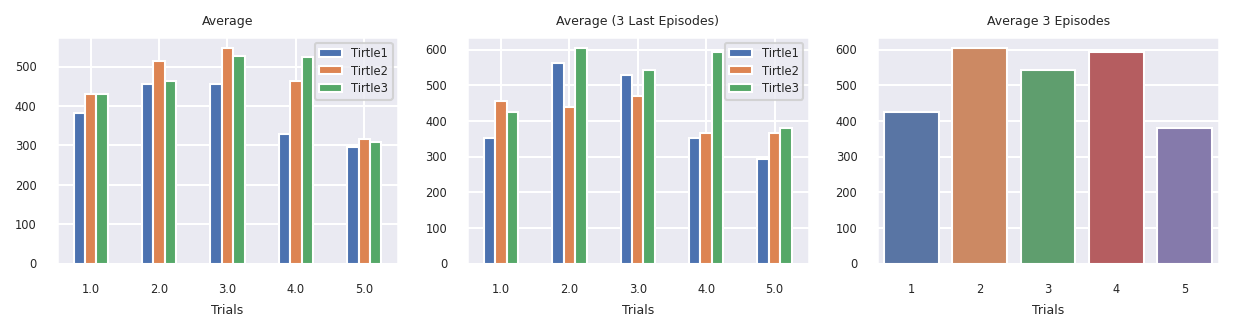}
         \caption{Single Frame}
         \label{fig:results_beam_rider_single_2}
     \end{subfigure}
     \begin{subfigure}[b]{1.0\textwidth}
         \centering
         \includegraphics[width=\textwidth]{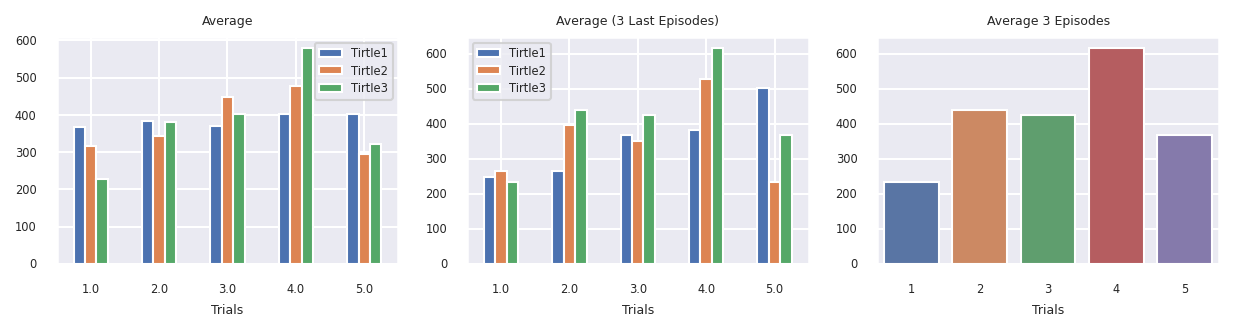}
         \caption{Stacked Frames}
         \label{fig:results_beam_rider_staked_2}
     \end{subfigure}
        \caption{Detailed results for Beam Rider}
        \label{fig:results_beam_rider_detailed}
\end{figure}

\begin{figure}[]
     \centering
     \begin{subfigure}[b]{1.0\textwidth}
         \centering
         \includegraphics[width=\textwidth]{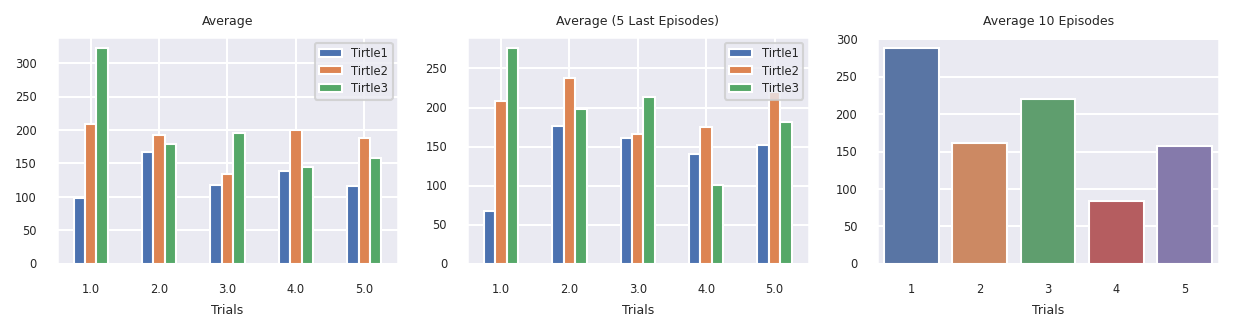}
         \caption{Single Frame}
         \label{fig:results_sapace_single_2}
     \end{subfigure}
     \begin{subfigure}[b]{1.0\textwidth}
         \centering
         \includegraphics[width=\textwidth]{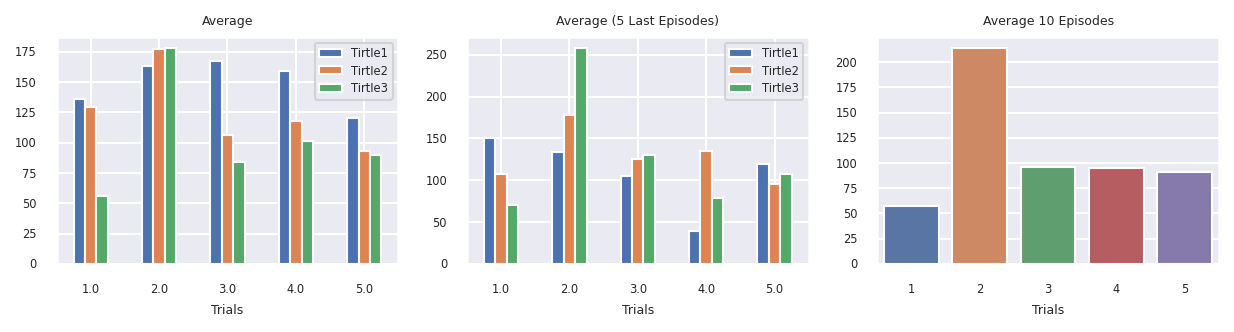}
         \caption{Stacked Frames}
         \label{fig:results_space_staked_2}
     \end{subfigure}
        \caption{Detailed results for Space Invaders}
        \label{fig:results_space_detailed}
\end{figure}

In Figures~\ref{fig:results_freeway_detailed},~\ref{fig:results_asteroids_detailed},~\ref{fig:results_battle_detailed},~\ref{fig:results_video_detailed},~\ref{fig:results_seaquest_detailed},~\ref{fig:results_asterix_detailed},~\ref{fig:results_beam_rider_detailed}, and~\ref{fig:results_space_detailed} we present the detailed results of \methodACRON\ for five runs on games Freeway, Asteroids, Battle Zone, Seaquest, Asterix, Beam Rider, and Space Invaders with 100 thousand frames. From left to right, we compare: (i)~the average reward of each tertile on each training trial; (ii)~the average reward of the last episodes of each tertile; (iii)~the average reward of the last training episodes of each trial. Full results are available at the COMPER's research repository\footnote{https://github.com/DanielEugenioNeves/COMPER-RELEASE-RESULTS}, which contains the training log files for implementations of COMPER and DQN agents and the Jupyter Notebooks used for analysis. The same graphs that we have presented here for COMPER are also available for DQN, as well as the learning curves and the minimum and maximum rewards for both the agents in each game.

In Freeway, the agent trained with \methodACRON\ remained without receiving rewards throughout many interactions until the end of the second training episode. However, once an agent received the first point, it began to accumulate rewards continuously from episode to episode (see~\ref{table:results_tertile_comparing} and Figure~\ref{fig:results_freeway_detailed}). The agents trained with DQN were somehow unable to get some score with only 100 thousand frames. For the game Asteroids, the agents got high scores in the first episodes and kept values close to the average until the end of the trials when they were only exploiting, and \methodACRON\ obtained the best results even at the end of each tertile.  For Battle Zone, \methodACRON\ did not reduce the average scores at the end of each tertile and maintained the best results compared to DQN (for both 100 thousand and 10 million frames). Video Pinball presented a lesser stable behavior for both \methodACRON\ and DQN, but \methodACRON\ obtained better average scores at the end of the trials and explicit learning progress at trial 1 with single frames and trial 5 with stacked frames (Figures ~\ref{fig:results_video_single_2} and~\ref{fig:results_video_staked_2}). For the game Seaquest, \methodACRON\ presented increasing scores from the first to the third tertile at all the trials, both with single and stacked frames and obtained the best result at the last tertile. For Asterix, the scores of \methodACRON\ did not decay at the end of the tertiles and we highlight the third trial with single frame (Figure~\ref{fig:results_asterix_single_2}) and the second trial with stacked frames (Figure~\ref{fig:results_asterix_staked_2}). Moreover, COMPER reached the last episodes with better results than DQN. For the Beam Rider, COMPER and DQN obtained very close average values and similar behaviors throughout the learning process. Except for the first trial, \methodACRON\ presented clear increasing scores at the end of the last episodes from the first to the third tertile, both with single frame and stacked frames. Finally, the same does not happen for the Space Invaders, and \methodACRON\ and DQN obtained close average scores and similar behaviors throughout the two first tertile, but \methodACRON\ got worse results than the DQN for the last tertile.

\section{Conclusion and Future Work}\label{sec:conclusion}

We propose a new method named \method\ (\methodACRON) to address relevant issues related to reinforcement learning concerning temporal credit assignment and experience replay. We assessed its effectiveness through its training and evaluation on ALE. The experimental results demonstrate that it is a promising method in that it achieves better performance (or at least equivalent) on difficult games, with considerably fewer observations of the environment. We highlight that \methodACRON\ scores better when using single frames instead of stacked frames and that the training runs with single frames were the ones that produced more similar transitions. Moreover, the size of the memory from which the agent sampled transitions is considerably lesser than used in the literature. That indicates that:  (i) our hypotheses are probably correct; (ii) the agent can learn from smaller memories; and (iii) similar transitions' sets capture the dynamics of the transitions regarding the long-term expected rewards.

According to~\citet{Lin1992}, the process of experiences replay can be more effective if the temporal difference error is determined by the discrepancy between multiple and consecutive predictions and not just two. Therefore, the way \methodACRON\ applies a recurrent neural network over the transitions set to approximate the target function seems to allow it to obtain better estimates. It appears to be associated with its use of similar transition' sets, which can produce a more effective update step for the action-value function. Moreover, algorithms based on AHC and Q-Learning become inefficient when they use the experiences obtained by trial-and-error only once to adjust the models and then discard that experiences, because that this can lead to the waste of rare experiences or expensive to obtain, such as those involving penalties. In this sense, the way \methodACRON\ uses its transition memories and generates a compact memory, from which it samples transitions to update the agent model, seems to increase its chances of observing relevant experiences again.

The main goal of this work was to present \methodACRON\ and the hypotheses that led us to propose it and evaluate its performance. However, there are still some aspects to investigate about this method, which is considerably different from other approaches based on temporal difference learning and experience replay. For instance, we chose the set of games we used because we consider them difficult and suitable to investigate our proposal. \methodACRON\ could be evaluated on other games, especially those that depend on learning a long-term action policy. Moreover, we have used strict similarity between transitions with the threshold hyperparameter equal to zero to Verify our hypotheses about the possibility of identifying similar transitions sets and use this with an RNN in a way to lead to more effectively Q-target predictions and to learn its transitions dynamics. Therefore, new experiments can be carried out by applying variations in the distance (or similarity) threshold value.

The main contribution of this work is to bring a new discussion about how the agent experiences can contribute to accelerating its learning and not only about how to deal with the retrieval of an agent’s experiences but also about how one can store them.

\acks{This study was financed in part by the Coordenação de Aperfeiçoamento de Pessoal de Nível Superior -- Brasil (CAPES) -- Finance Code 001. It has also received partial funding from Brazilian Conselho Nacional de Desenvolvimento Científico e Tecnológico (CNPq) and from Pontifical Catholic University of Minas Gerais (PUC Minas).}

\appendix

\section{Experimental Setup and Parameters}\label{sec_apendix:experimental_setup}

Experimental methodology and evaluation protocol were discussed in Section~\ref{sec:metodologia}. However, there are specific sets of hyperparameters for different purposes. Some of them are independent of the evaluated methods while others address aspects related to each of them, as shown in the following.

\subsection*{A.1. General Parameters of ALE and the Evaluation Protocol}\label{subsec:general_parameters}

\begin{table}[!b]
\centering 
\begin{tabular}{|l|l|l|}
\hline
\textbf{Hyperparameter} & \textbf{Value} & \textbf{Description} \\ \hline
Action set & Full & 18 actions are always available to the agent. \\ \hline
Frame skip & 5 & Each action lasts 5 time steps. \\ \hline
Stochasticity ($\varsigma$) & 0.25 & We used $\varsigma$ = 0.25 for all experiments. \\ \hline
Lives signal used & False & \begin{tabular}[c]{@{}l@{}}We did not use the game-specific information\\ about the number of lives the agent has at each\\ time step.\end{tabular} \\ \hline
\begin{tabular}[c]{@{}l@{}}Number of episodes\\ used for evaluation\end{tabular} & 3 or 10 & \begin{tabular}[c]{@{}l@{}}We report the average score obtained in the last\\ 3 or 10 episodes used for learning. See Subsection\\ \ref{sec:experimental_procedures} for details.\end{tabular} \\ \hline
\begin{tabular}[c]{@{}l@{}}Number of episodes \\ used for evaluation in\\ the continual learning\end{tabular} & 3 or 5 & \begin{tabular}[c]{@{}l@{}}In the continual learning setting, we report the \\ average score obtained in the last 3 or 5 episodes\\ at the end of each tertile. See Subsection~\ref{sec:experimental_procedures} for \\details.\end{tabular} \\ \hline
\begin{tabular}[c]{@{}l@{}}Number of frames\\ used for learning ($T$)\end{tabular} & 100,000 & \begin{tabular}[c]{@{}l@{}}We report the scores after 100,000 frames and\\ at the end of each tertile of episodes.\end{tabular} \\ \hline
Number of trials & 5 & \methodACRON\ and DQN were evaluated in 5 trials. \\ \hline
\end{tabular}
\caption{Parameters used to set up the environment and evaluate the methods.}
\label{table:ale_parameters}
\end{table}

In Table~\ref{table:ale_parameters}, we present the parameters used to set up the Arcade Learning Environment (ALE), to evaluate the agent's learning, and to summarize the results. The other parameters used to configure the environment have the same values define by~\citet{machado18arcade}. Those that refer to summarizing the results were defined as discussed in Subsection~\ref{sec:experimental_procedures}.
In Table~\ref{table:exploration_parameters}, we present the parameters related to the exploration (and exploitation) rate, and the discount factor.
\begin{table}[]
\centering
\begin{tabular}{|l|l|l|}
\hline
\textbf{Hyperparameter} & \textbf{Value} & \textbf{Description} \\ \hline
Discount factor ($\gamma$) & 0.99 & \begin{tabular}[c]{@{}l@{}}The discount factor applied in the update rule on\\ the expected long-term reward.\end{tabular} \\ \hline
Initial exploration rate ($\epsilon$) & 1.0 & \begin{tabular}[c]{@{}l@{}}Probability of a random action will be taken at\\ each time step.\end{tabular} \\ \hline
Final exploration rate ($\epsilon$) & 0.001 & \begin{tabular}[c]{@{}l@{}}Probability of a random action will be taken at\\ each time step.\end{tabular} \\ \hline
Final exploration frame & 90,000 & \begin{tabular}[c]{@{}l@{}}Number of frames over which $\epsilon$ is linearly\\ annealed.\end{tabular} \\ \hline
\end{tabular}
\caption{Discount factor and exploration rate used by \methodACRON\ and DQN.}
\label{table:exploration_parameters}
\end{table}

\subsection*{A.2. Specific parameters of \methodACRON}\label{subsec:specific_comper_parameters}

As discussed in Section~\ref{sec:comper}, \methodACRON\ uses different models and architectures of deep neural networks to approximate their value functions, two transition memories, and a reduction process from one to another, requiring specific sets of parameters, which are presented in Tables~\ref{table:comper_cnn_parameters}, \ref{table:comper_lstm_parameters}, and~\ref{table:comper_QLSTM_parameters}.

\begin{table}[ht]
\centering
\begin{tabular}{|l|l|l|}
\hline
\textbf{Hyperparameter} & \textbf{Value} & \textbf{Description} \\ \hline
Step-size ($\alpha$) & 0.00025 & Step-size used by RMSProp. \\ \hline
Gradient momentum & 0.95 & Gradient momentum used by RMSProp. \\ \hline
\begin{tabular}[c]{@{}l@{}}Squared gradient\\ momentum\end{tabular} & 0.95 & \begin{tabular}[c]{@{}l@{}}Squared gradient (denominator) momentum used by\\RMSProp.\end{tabular} \\ \hline
Min squared gradient & 0.01 & \begin{tabular}[c]{@{}l@{}}Constant added to the denominator of the RMSProp\\ update.\end{tabular} \\ \hline
\end{tabular}
\caption{Parameters used in CNN training by \methodACRON.}
\label{table:comper_cnn_parameters}
\end{table}

\begin{table}[ht]
\centering
\begin{tabular}{|l|l|l|}
\hline
\textbf{Hyperparameter} & \textbf{Value} & \textbf{Description} \\ \hline
Step-size ($\alpha$) & 0.00025 & Step-size used by RMSProp. \\ \hline
Gradient momentum & 0 & Gradient momentum used by RMSProp. \\ \hline
Decay & 0.9 & \begin{tabular}[c]{@{}l@{}}Discounting factor for the history and coming\\ gradient used by RMSProp.\end{tabular} \\ \hline
Epsilon & 1e-10 & \begin{tabular}[c]{@{}l@{}}Constant added to the denominator of the RMSProp\\ update.\end{tabular} \\ \hline
\end{tabular}
\caption{Parameters used in LSTM training.}
\label{table:comper_lstm_parameters}
\end{table}

\begin{table}[ht]
\centering
\begin{tabular}{|l|l|l|}
\hline
\textbf{Hyperparameter} & \textbf{Value} & \textbf{Description} \\ \hline
\begin{tabular}[c]{@{}l@{}}Transition memory\\ size\end{tabular} & \begin{tabular}[c]{@{}l@{}}100,000\end{tabular} & \begin{tabular}[c]{@{}l@{}}The transition memory size is defined to stores up to\\ 100,000 sets of last similar transitions. However, it\\ is reduced every when these sets are sampled.\end{tabular} \\ \hline

Replay start size & 100 & \begin{tabular}[c]{@{}l@{}}Number of frames over which a random policy is \\ executed to first populate the transition memory.\end{tabular} \\ \hline
\begin{tabular}[c]{@{}l@{}}History length\\ (single frame) \end{tabular}& 1 & \begin{tabular}[c]{@{}l@{}}Number of most recent frames the agent observed\\ that are given as input to the CNN when using\\ single frame.\end{tabular} \\ \hline
\begin{tabular}[c]{@{}l@{}}History length\\ (stacked frames) \end{tabular}& 4 & \begin{tabular}[c]{@{}l@{}}Number of most recent frames the agent observed\\ that are given as input to the CNN when using\\ stacked frames.\end{tabular} \\ \hline
\begin{tabular}[c]{@{}l@{}}Train Frequency ($TF$)\\ \end{tabular}& 4 & \begin{tabular}[c]{@{}l@{}}Number of iterations between the CNN model\\ updates.\end{tabular} \\ \hline

\begin{tabular}[c]{@{}l@{}}Transitions batch size \\($K$)  \end{tabular}& 32 & \begin{tabular}[c]{@{}l@{}}The number of transitions sampled from the\\ reduced transitions memory to update the CNN.\end{tabular} \\ \hline
\begin{tabular}[c]{@{}l@{}}Similar transitions\\ batch size\end{tabular} & 1,000 & \begin{tabular}[c]{@{}l@{}}The number of sets of similar transitions taken\\ from the transition memories to produce the\\ reduced transition memory.\end{tabular} \\ \hline
\begin{tabular}[c]{@{}l@{}}Similarity Threshold\\${\delta}$ \end{tabular} & 0.0 & \begin{tabular}[c]{@{}l@{}}Maximum distance to consider two transitions as similar\end{tabular} \\ \hline
\begin{tabular}[c]{@{}l@{}}Update Frequency\\ ($UTF$)\end{tabular} & 100 & \begin{tabular}[c]{@{}l@{}}The frequency with which the target network \\(LSTM) is trained and the reduced transition \\memory is updated by the QLSTM algorithm.\end{tabular} \\ \hline
Color averaging & True & \begin{tabular}[c]{@{}l@{}}The observation received is the average between\\ the previous and the current frame.\end{tabular} \\ \hline
\begin{tabular}[c]{@{}l@{}}Number of different\\ colors\end{tabular} & 8 & \begin{tabular}[c]{@{}l@{}}NTSC is the color palette in which each screen is \\ encoded but only the luminance channel is used\end{tabular} \\ \hline
\end{tabular}
\caption{General parameters used by \methodACRON.}
\label{table:comper_QLSTM_parameters}
\end{table}



\subsection*{A.3. Specific parameters of DQN}\label{subsec:specific_dqn_parameters}

In Table~\ref{table:dqn_parameters}, we present the parameter used on DQN. Those related to the memory of experiences and the target network update frequency (\textbf{in bold}) were adjusted due to the smaller number of frames used in the experiments. All others have the same values defined by~\citet{machado18arcade}. 

\begin{table}[h]
\centering
\begin{tabular}{|l|l|l|}
\hline
\textbf{Hyperparameter} & \textbf{Value} & \textbf{Description} \\ \hline
Step-size ($\alpha$) & 0.00025 & Step-size used by RMSProp. \\ \hline
Gradient momentum & 0.95 & Gradient momentum used by RMSProp. \\ \hline
\begin{tabular}[c]{@{}l@{}}Squared gradient\\ momentum\end{tabular} & 0.95 & \begin{tabular}[c]{@{}l@{}}Squared gradient (denominator) momentum\\ used by RMSProp.\end{tabular} \\ \hline
Min squared gradient & 0.01 & \begin{tabular}[c]{@{}l@{}}Constant added to the denominator of the\\ RMSProp update.\end{tabular} \\ \hline
\textbf{Replay memory size} & 100,000 & \begin{tabular}[c]{@{}l@{}}The samples used in the algorithm’s updates\\are drawn from the last 100 thousand recent\\ frames.\end{tabular} \\ \hline
\textbf{Replay start size} & 1,000 & \begin{tabular}[c]{@{}l@{}}Number of frames over which a random policy\\ is executed to first populate the replay\\ memory.\end{tabular} \\ \hline
\begin{tabular}[c]{@{}l@{}}History length\\ (stacked frames) \end{tabular}& 4 & \begin{tabular}[c]{@{}l@{}}Number of most recent frames the agent observed\\ that are given as input to the CNN.\end{tabular} \\ \hline
\begin{tabular}[c]{@{}l@{}}Update frequency\end{tabular} & 4 & \begin{tabular}[c]{@{}l@{}}Number of actions the agent selects between\\ successive updates.\end{tabular} \\ \hline

\begin{tabular}[c]{@{}l@{}}Update frequency of\\ target network\end{tabular} & 1,000 & \begin{tabular}[c]{@{}l@{}}Number of actions the agent selects between\\ successive updates of the target CNN parameters.\end{tabular} \\ \hline

Frame pooling & True & \begin{tabular}[c]{@{}l@{}}The observation received consists of the\\ maximum pixel value between the previous\\ and the current frame.\end{tabular} \\ \hline
\begin{tabular}[c]{@{}l@{}}Number of different\\ colors\end{tabular} & 8 & \begin{tabular}[c]{@{}l@{}}NTSC is the color palette in which each\\ screen is encoded but only the luminance\\ channel is used.\end{tabular} \\ \hline
\end{tabular}
\caption{Parameters used by DQN}
\label{table:dqn_parameters}
\end{table}


\section*{Appendix B. Deep Neural Network Architectures used in \methodACRON}

The convolutional neural network (CNN) used in \methodACRON\ to approximate the Q-Value function has the same general architecture as proposed by \citet{mnih2015humanlevel} and used in \citet{machado18arcade}. The difference is in the input layer because when \methodACRON\ is using a single frame it represents the frames as a matrix of $84 \times 84 \times 1$ and when using stacked frames its shape is $84 \times 84 \times 4$. This representation is produced by a preprocessing step that reduces the original dimensionality of the game frames and extracts the luminance channel. Another difference is in the output layer. As recommended by~\citet{machado18arcade} and discussed in Section~\ref{sec:experimental_procedures}, the \methodACRON\ agent has access to the complete set of actions on ALE, independently from the game, so the output layer is a fully-connected layer with a single output for each of the 18 valid actions. The hidden layers are three convolutional and one fully-connected. The first convolutional layer applies 32 filters of $8 \times 8$ with stride 4 on the input frame image representation, followed by a ReLU nonlinearity. The second layer convolves 64 filters of $4 \times 4$ with stride 2 and also applies a ReLU nonlinearity. The third hidden layer convolves 64 filters of $3 \times 3$ with stride 1, again followed by a ReLU nonlinearity. The final hidden layer consists of 512 units. The RMSProp optimizer uses the parameters shown in Table~\ref{table:comper_cnn_parameters} and this architecture is detailed in Table~\ref{table:cnn_architecture}.

\begin{table}[t]
\centering
\begin{tabular}{@{}|l|l|l|l|@{}}
\hline
\textbf{Layer (type)} & \textbf{Definition} & \textbf{Output Shape} & \textbf{Parameters} \\ \hline
Conv1 (Conv2D) & \begin{tabular}[c]{@{}l@{}}Filters = (32, (8, 8))\\ Strides= (4, 4)\\ Input = (84, 84, 1)\textsuperscript{*} or\\Input = (84, 84, 4)\textsuperscript{**}\\\end{tabular} & (None, 20, 20, 32) & \begin{tabular}[c]{@{}l@{}}2,080\textsuperscript{*}\\ or\\ 8,192\textsuperscript{**}\end{tabular} \\ \hline
Conv2 (Conv2D) & \begin{tabular}[c]{@{}l@{}}Filters = (64, (4, 4))\\ Strides= (2, 2)\end{tabular} & (None, 9, 9, 64) & 32,832 \\ \hline
Conv3 (Conv2D) & \begin{tabular}[c]{@{}l@{}}Filters = (64, (3, 3))\\ Strides= (1, 1)\end{tabular} & (None, 7, 7, 64) & 36,928 \\ \hline
Flatten & Flatten & (None, 3136) & 0 \\ \hline
Dense1 (Dense) & 512 Units & (None, 512) & 1,606,144 \\ \hline
Dense2 (Dense) & 18 Units & (None, 18) & 9,234 \\ \hline
\end{tabular}
\caption{CNN architecture used in \methodACRON\ with single frame. Note that \textsuperscript{*} indicates the values for single frame and \textsuperscript{**} the values for stacked frames.}
\label{table:cnn_architecture}
\end{table}

The LSTM network that predicts values for the target function has the input layer with shapes of $1 \times 14114$ or $1 \times 56450$, which represents a transition $\tau$ for single frame and stacked frames, respectively. The batch size is 16. The output is a fully-connected linear layer with only a single output for the predicted Q-Value to the input transition. The hidden layers are three LSTM and one fully connected.  The first LSTM layer contains 64 internal units, applies a hyperbolic tangent activation function and sequence return. The second layer contains 32 internal units, and also applies a hyperbolic tangent activation function and sequence return. The third LSTM layer also contains 32 internal units and applies a hyperbolic tangent activation function, but has no sequence return. The last hidden layer contains 18 fully connected units and applies a ReLU nonlinearity. The RMSProp optimizer uses the parameters shown in Table~\ref{table:comper_lstm_parameters} and this architecture is detailed in Table~\ref{table:lstm_architecture}. 

\begin{table}[!b]
\centering
\begin{tabular}{@{}|l|l|l|l|@{}}
\hline
\textbf{Layer (type)} & \textbf{Definition} & \textbf{Output Shape} & \textbf{Parameters} \\ \hline
Lstm1 (LSTM) & \begin{tabular}[c]{@{}l@{}}Units = 64\\ Return Sequences = True\\ Input = (1, 14114)\textsuperscript{*} or\\ Input = (1, 56450)\textsuperscript{**}\end{tabular} & (None, 1, 64) & \begin{tabular}[c]{@{}l@{}}3,629,824\textsuperscript{*} or\\14,467,840\textsuperscript{**}\end{tabular} \\ \hline
Lstm2 (LSTM) & \begin{tabular}[c]{@{}l@{}}Units = 32\\ Return Sequences = True\end{tabular} & (None, 1, 32) & 12,416 \\ \hline
Lstm3 (LSTM) & Units = 32 & (None, 32) & 8,320 \\ \hline
Dense1 (Dense) & Units = 18 & (None, 18) & 594 \\ \hline
Dense2 (Dense) & Units = 1 & (None, 1) & 19 \\ \hline
\end{tabular}
\caption{LSTM architecture used in \methodACRON. Note that \textsuperscript{*} indicates the values for single frame and \textsuperscript{**} the values for stacked frames.}
\label{table:lstm_architecture}
\end{table}

\section*{Appendix C. Relevant Aspects of Agent Evaluation in ALE}\label{sec_apendix:ale_challenges_important_aspects}

\citet{bellemare13arcade} proposed the Arcade Learning Environment (ALE) that consists of an Atari 2600 game emulator and a set of challenges that includes non-determinism, stochasticity, and environments with different values and ranges for awarding rewards, allowing the development and evaluation of reinforcement learning agents. \citet{machado18arcade} presented a revision of ALE and introduced features such as the support for different difficulty game levels. They also discussed the methodological differences between relevant works in the literature regarding the agent evaluation procedure, which makes the analysis and comparison of results difficult. Therefore, the authors proposed a very well-defined methodology for agents evaluation using ALE and presented a benchmarking with the experimental results for DQN~\citep{mnih2015humanlevel} and SARSA(\(\lambda\))\(+\) blob-PROST~\citep{aamas2016Liang}.

An agent should interact with as many games as possible without using any specific information from these games, acting based just on its perceptions from the video streaming, as these games present multiple and different tasks, are challenging even for human beings, and are free of possible biases introduced by researchers. This interaction with the environment happens episodically, and each episode starts in the initial game state and finishes when the game is over at a final state, as when the agent runs out of all its attempts, exceeds a time limit, or reaches its goal. The undiscounted sum of the agent rewards at each episode is the principal measure of its performance, but this can make it difficult to obtain more detailed information about its learning progress. The agent may have learned to focus on actions that lead to subsequent and small rewards, or it may have simply learned to survive in the game by increasing its chances of receiving some reward. Even so, besides being a common main objective of reinforcement learning agents, the evaluation of its total rewards allows to verify if a method leads the agent to an effective learning process, how stable is its behavior between different training trials, and how the different methods affect that evaluation considering the distinct challenges represented by the diversity of games~\citep{machado18arcade}.

In most games, an agent has a finite number of \say{lives}, which are lost one by one each time it fails to fulfill its goal, causing the game to ends when the available number of lives runs out. On the other hand, the success of an agent also leads to the game ends. Many factors can cause the game to ends: a time limit that ends regardless of the number of lives lost or an agent decision that takes you to a point in the game from which there is no way out. In ALE, it is possible to end a training episode as soon as an agent loses its first life or only when the game effectively ends. The second option can prevent an agent from learning the meaning of losing a life, but ending the game right after the agent loses its first life can make it just learns not to die rather than accumulate rewards. This way, while both approaches are present in the literature, ending an episode when an agent loses a life is detrimental to its performance. Thinking about the main objective of an agent, (i.e. learning a policy that leads to the highest discounted accumulated reward in the long run) learning \say{just} not to die may not lead to the best choice of actions. Therefore, \citet{machado18arcade} strongly recommend that only the game is over signal be used to end an episode. Therefore,the differences in the literature regarding the results measurement and the agent's performance evaluation based on these results imply a relevant methodological decision about when each training episode ends.

The way results are measured and summarized can hamper or improve the analysis of an agent learning process. When evaluating an agent on a small set of problems, it is possible to describe its learning curves in detail, measuring important information such as its learning speed, its best-presented performance, the stability of its algorithm, and how much its performance improves while it observes more and more data. But this process becomes difficult on the set of all games present in ALE. To facilitate comparative analysis, many authors present their results by numerically summarizing the agents' performance for each game, but with different approaches, and it impairs the direct comparison of results. To~\citet{machado18arcade}, the analysis of the results should be carried out on a fixed number of the last episodes at different times of agent training, allowing the evaluation of its learning evolution. In this way, unstable methods (that show a great deal of variation on the reward values between different episodes) will present a low evaluation compared to those that improve steadily and continuously. Although the performance at the end of the training is of most interest, this approach allows one to evaluate agent evolution through stated points without tracing the entire learning curve, allowing researchers to report results earlier during experiments, minimizing the problem of agents' evaluation associated with the computational cost. In this sense, and according to the authors, agents should be evaluated over the training data and not in \say{test mode} because we are interested in evaluating its learning and not its in-game performance.

Training an agent over a given number of episodes before evaluating it can yield misleading results, as the duration of each episode can vary widely between different games. Furthermore, the better or worse the agent's performance, the longer or shorter the training episodes could be for the same game in distinct runs. Thus, agents that learn good policies earlier tend to receive more training data than those that learn more slowly, significantly increasing the magnitude of the difference between them for a specific number of episodes. Therefore, a more interesting approach is to measure the amount of training data in terms of the total number of observed frames by an agent, which facilitates results reproducibility, analysis, and comparison. Even so, interrupting an ongoing episode as soon as the total number of frames is reached or waiting for the end of this episode does not represent a clear choice in the literature. So,~\citet{machado18arcade} recommended ending the agent training based on a total number of frames without interrupting the episode in which this total is reached.

Non-determinism and stochasticity consist of two fundamental aspects of reinforcement learning since an agent needs to learn to deal with partially observable and dynamic environments. Therefore, \citet{machado18arcade} introduced a form of stochasticity named \say{stick actions} that consists of repeating the last action performed by an agent with a given probability (ignoring its current choice), and an approach named \say{frame skipping} that consists of limiting the agent's ability to decide on the behavior of the environment through the repetition of its last action on the next frames. Both are defined by specific parameters of the ALE. 

Regarding environment states representation, three main approaches are present in the literature: (i) \textit{color averaging}, (ii) \textit{frame pooling}, and (iii) \textit{frame stacking}. The first two consist of composing a representation of successive frames in a single one to reduce visual issues resulting from the hardware limitation of Atari 2600, and ALE implements the first in a parametric way. In turn, frame stacking consists of concatenating a certain number of frames already observed with the most recent frame, seeking to obtain a more informative observation space for the agent and making it easier to infer trajectories of moving objects in the game environment. It is not implemented directly by ALE and is an algorithm decision but. According to \citet{machado18arcade}, color averaging and frame pooling may end up removing the most interesting form of partial observation in ALE, and frame stacking  reduces the degree of partial observability.

Finally,~\citet{machado18arcade} also pointed out divergences in the literature regarding the processes of finding out the values of hyperparameters. Some research works use all of the games, while others make specific adjustments for each game. To evaluate the agents' generalization capacity, the authors recommended splitting the set of games analogously to supervised learning. Therefore, one should only use a smaller subset of games to adjust the values of the hyperparameters and then use those values to train the agent, without further adjustment, across the entire game set.

\vskip 0.2in
\bibliography{comper}
\bibliographystyle{plainnat}

\end{document}